\documentclass[11pt]{article}

\usepackage{acl}

\usepackage{times}
\usepackage{latexsym}
\usepackage[T1]{fontenc}
\usepackage[utf8]{inputenc}
\usepackage{microtype}

\usepackage{amsmath,mathtools}
\usepackage{amssymb}
\usepackage{booktabs}
\usepackage{multirow}
\usepackage{enumitem}
\usepackage{placeins}
\usepackage{needspace}
\usepackage{xcolor}
\usepackage{xspace}
\usepackage{algorithm}
\usepackage[noend]{algpseudocode}
\usepackage{framed}
\usepackage{listings}
\usepackage{graphicx}
\usepackage{cuted}
\usepackage{url}
\usepackage{hyperref}

\providecommand{\Description}[1]{}

\usepackage[most]{tcolorbox}
\newtcolorbox{casecard}[1]{%
  enhanced,
  colback=white, colframe=black!70,
  boxrule=0.6pt, arc=1.5mm,
  left=5pt, right=5pt, top=4pt, bottom=4pt,
  title=\textbf{#1}, fonttitle=\bfseries\small
}

\definecolor{shadecolor}{rgb}{0.92,0.92,0.92}

\definecolor{promptbg}{RGB}{242,247,255}
\definecolor{promptborder}{RGB}{52,95,175}
\newtcolorbox{promptcard}[1]{%
  enhanced, breakable,
  colback=promptbg, colframe=promptborder,
  boxrule=0.6pt, arc=1mm,
  left=6pt, right=6pt, top=4pt, bottom=4pt,
  title={\textbf{#1}}, fonttitle=\bfseries\small,
  fontupper=\ttfamily\small\raggedright,
}
\usepackage{tikz}
\usetikzlibrary{positioning,arrows.meta,calc,fit,matrix,shapes,shapes.geometric}
\usepackage{pgfplots}
\pgfplotsset{compat=1.18}

\newcommand{\bench}{FinToolBench}
\newcommand{\methodname}{FATR}
\newcommand{\tir}{\textsc{TIR}\xspace}
\newcommand{\tesr}{\textsc{TESR}\xspace}
\newcommand{\cer}{\textsc{CER}\xspace}
\newcommand{\soft}{\textsc{Soft Score}\xspace}
\newcommand{\css}{\textsc{CSS}\xspace}
\newcommand{\tmr}{\textsc{TMR}\xspace}
\newcommand{\imr}{\textsc{IMR}\xspace}
\newcommand{\dmr}{\textsc{DMR}\xspace}

\newcommand{\realt}{\texttt{realtime}}
\newcommand{\dailyt}{\texttt{daily}}
\newcommand{\asfiledt}{\texttt{as\_filed}}
\newcommand{\periodict}{\texttt{periodic}}
\newcommand{\statict}{\texttt{static}}

\newcommand{\informational}{\texttt{informational}}
\newcommand{\advisory}{\texttt{advisory}}
\newcommand{\transactional}{\texttt{transactional}}

\title{FinToolBench: Evaluating LLM Agents for Real-World Financial Tool Use}

\author{
  \textbf{Jiaxuan Lu}$^{1,*}$ \quad \textbf{Kong Wang}$^{2,*}$ \quad \textbf{Yemin Wang}$^{3}$ \quad
  \textbf{Qingmei Tang}$^{4}$ \quad \textbf{Hongwei Zeng}$^{5}$ \quad \textbf{Xiang Chen}$^{6}$ \\
  \textbf{Jiahao Pi}$^{1}$ \quad \textbf{Shujian Deng}$^{1}$ \quad \textbf{Lingzhi Chen}$^{1}$ \quad
  \textbf{Yi Fu}$^{1}$ \quad \textbf{Kehua Yang}$^{2,\dagger}$ \quad \textbf{Xiao Sun}$^{1,\dagger}$ \\
  $^{1}$Shanghai AI Laboratory \quad $^{2}$Hunan University \quad
  $^{3}$Xiamen University \\
  $^{4}$Tencent \quad $^{5}$University of Chinese Academy of Sciences \quad $^{6}$Tongji University \\
  $^{*}$Equal contribution \quad $^{\dagger}$Corresponding authors
}

\begin{document}
\maketitle

\begin{abstract}
Large Language Model (LLM) agents are increasingly used as interfaces to financial data, yet existing evaluations often score final answers while leaving tool traces weakly tested.
This is risky in finance: a valid-looking call can still be unacceptable if it uses stale data, escalates user intent, or crosses market and regulatory domains.
We introduce \textbf{\bench}, a runnable benchmark of 760 real free-tier financial tools paired with 295 tool-required questions (166 single-tool, 129 multi-tool).
Each tool is annotated with finance attributes, \textit{i.e.}, timeliness, intent type, and regulatory domain, which support call-level compliance metrics (\tmr, \imr, \dmr) computed directly from execution traces.
We also provide \textbf{FATR} (Finance-Aware Tool Routing), a lightweight reference baseline that retrieves candidate tools, injects finance attributes into tool cards, and records auditable traces.
Across seven LLM backends, including Doubao-Seed-1.6, Claude-Sonnet-4.5, Grok-3-beta, and Gemini-3.1-Pro, no single model dominates capability and compliance jointly, separating aggressive callers from precise but conservative ones.
Code and data: \url{https://github.com/Double-wk/FinToolBench}.
\end{abstract}

\begin{figure}[t]
\centering
\includegraphics[width=\columnwidth]{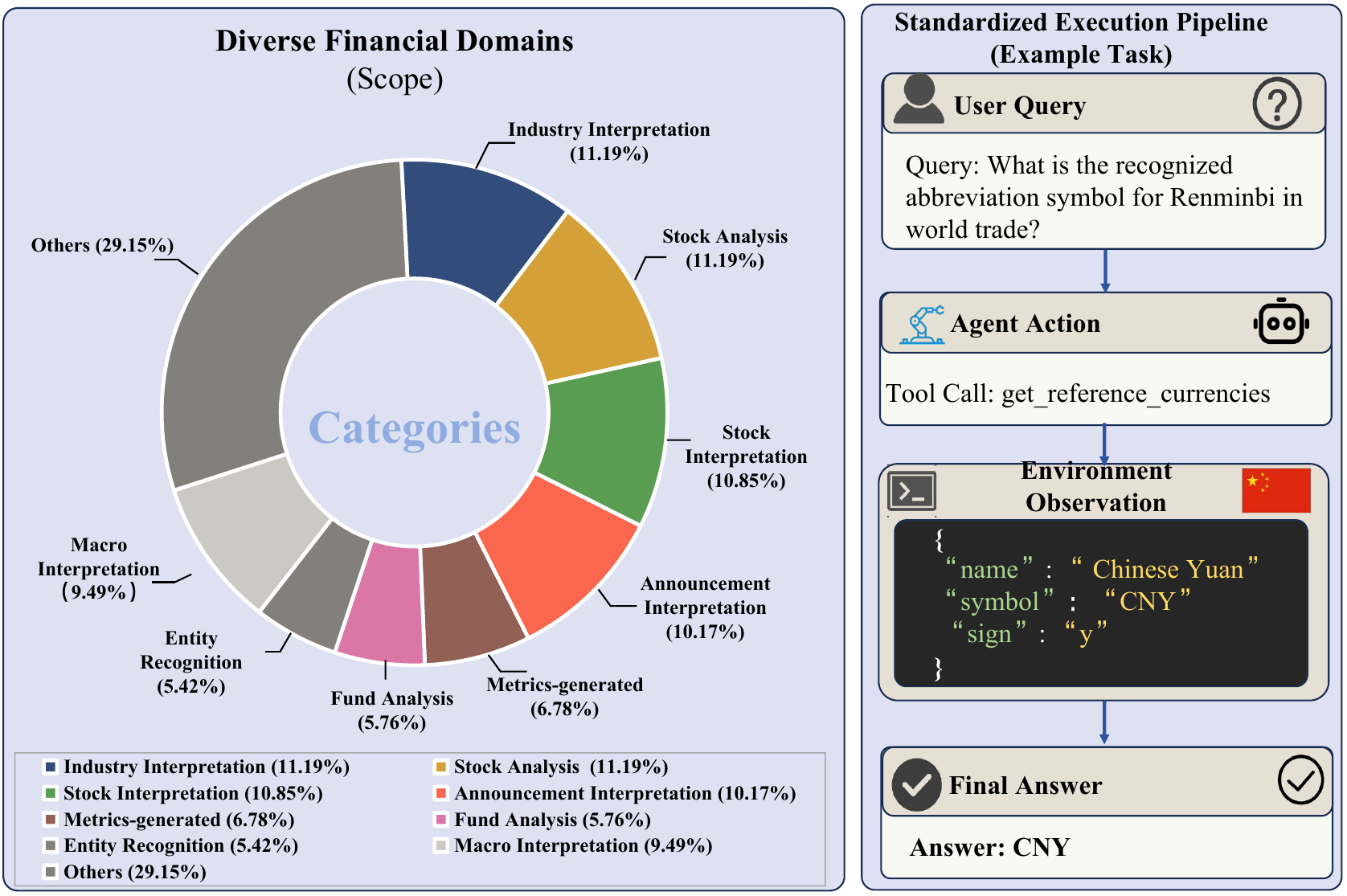}
\caption{FinToolBench overview. Left: the scope of our benchmark across representative categories. Right: an example of the standardized execution pipeline, where an LLM agent selects a tool, observes the environment output, and produces a final answer with an auditable tool trace.}
\label{fig:overview}
\end{figure}

\section{Introduction}
Large Language Models (LLMs) are moving financial analysis from static question answering toward dynamic interaction with APIs, databases, and computational tools.
In this setting, the tool trace is part of the answer.
A response may appear grounded because it contains tool outputs, yet still be unreliable if the agent retrieved stale data, called a drifting endpoint, or used a tool from the wrong market domain~\citep{guo2024stabletoolbench}.
Evaluation must therefore assess not only whether tools are invoked and executed successfully, but also whether the resulting tool trace is acceptable under finance-specific constraints, especially as agents operate over longer horizons and tool use itself evolves~\citep{lu2026beyond,jiang2025scp}.

Existing benchmarks leave a gap between what is easy to measure and what is necessary to trust.
General tool benchmarks emphasize API correctness and executability~\citep{guo2024stabletoolbench} but rarely test finance-specific acceptability. Finance benchmarks focus on knowledge- or document-centric QA and involve virtually no executable tools, relying on static datasets or a negligible number of mock interfaces.
We argue that current metrics are blind to three recurring failure modes essential for financial reliability: (i) \textit{timeliness}---a question asking for ``current'' exchange rates is fundamentally unanswered if the agent retrieves a daily snapshot, even if the API call is syntactically perfect; (ii) \textit{intent restraint}---an agent must differentiate informational queries from transactional actions and never escalate to execution without explicit authorization; and (iii) \textit{domain alignment}---the chosen tool chain must adhere to the regulatory and market domain of the query (\textit{e.g.}, using equity tools for a cryptocurrency inquiry is a domain hallucination).

To address these gaps, we introduce \textbf{\bench}, a runnable benchmark built from real free-tier tools and tool-required questions.
\bench{} scales financial agent evaluation to 760 executable tools and 295 tool-required items (166 single-tool, 129 multi-tool).
Each tool is annotated with three finance attributes, \textit{i.e.}, timeliness, intent type, and regulatory domain, enabling us to compute call-level compliance mismatch rates (\tmr{}, \imr{}, \dmr{}) alongside standard invocation and execution metrics.
We further provide \textbf{FATR} (Finance-Aware Tool Routing), a lightweight reference baseline that retrieves a small candidate set, injects finance attributes into tool cards, and stabilizes execution with caching, retries, and output compression.
Figure~\ref{fig:overview} sketches the benchmark scope and the standardized execution pipeline.

In summary, this paper makes three contributions:
(1) \bench{}: a benchmark of 760 free-tier financial tools and 295 tool-required questions producing auditable tool traces under real execution.
(2) Finance-aware evaluation: capability metrics plus call-level compliance mismatch rates (\tmr{}, \imr{}, \dmr{}) measuring violations of timeliness, intent restraint, and domain alignment.
(3) FATR: a lightweight reference baseline and execution harness for evaluating finance-aware tool routing under a fixed stack.

\section{Related Work}\label{sec:related}
\subsection{Tool-Using Agents and Benchmarks}
Tool-augmented agents interleave reasoning with external actions to improve grounding and support up-to-date answers~\citep{yao2022synergizing,schick2023toolformer,patil2024gorilla,qin2024toolllm}.
Benchmarks evaluate tool selection and calling at scale (API-Bank~\citealp{li2023api}; StableToolBench~\citealp{guo2024stabletoolbench}) and long-horizon interaction in realistic environments~\citep{liu2024agentbench,mialon2024gaia,zhou2024webarena,drouin2024workarena,narasimhan2024tau}.
Recent efforts sharpen the focus toward tool-interface competence and agentic behavior, including BFCL~\citep{patil2025berkeley} and $\tau^2$-bench~\citep{barres2025tau2}, and study agents in settings where tools or capabilities evolve over time and long-horizon traces are central artifacts~\citep{lu2026beyond,jiang2025scp,wan2026deep,yang2025multi}.

\subsection{Financial Benchmarks and Evaluation}
In finance, most benchmarks emphasize domain knowledge and document-centric QA rather than executable tool use. Examples include FinanceBench~\citep{islam2023financebench}, OpenFinData~\citep{openfindata2024}, and report-focused datasets such as FinQA~\citep{chen2021finqa} and TAT-QA~\citep{zhu2021tat}. Recent works like FinEval~\citep{guo2025fineval}, FLAME~\citep{guo2025flame}, and the Finance Agent Benchmark~\citep{bigeard2025finance} broaden knowledge coverage, but none release a standardized large tool library or define call-level compliance metrics. Safety-oriented agent evaluations~\citep{xia2025safetoolbench,tur2025safearena} probe deliberate misuse but are not finance-specific and do not operationalize domain-grounded constraints like timeliness, intent limits, and regulatory scope. \bench{} addresses this by pairing a fully runnable tool inventory with tool-required questions and explicitly defining finance constraints at the level of each tool call via a lightweight, auditable attribute schema, enabling direct measurement of timeliness, intent, and domain mismatches from execution traces rather than relying solely on final-answer correctness or generic safety checks.

\begin{figure*}[t]
\centering
\includegraphics[width=\textwidth]{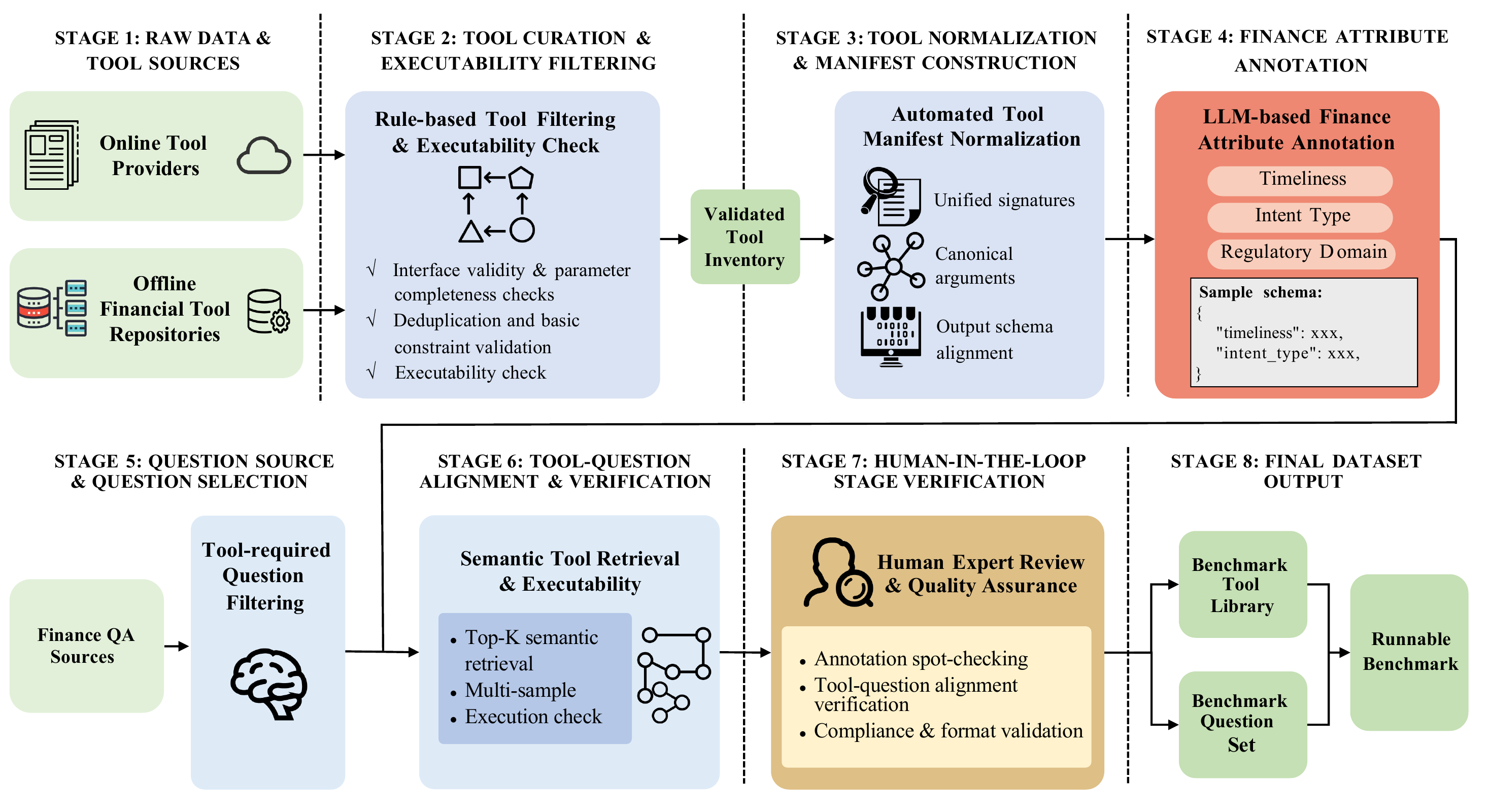}
\caption{FinToolBench dataset construction pipeline. Stage 1 collects raw tool sources. Stage 2 performs tool curation and executability filtering to obtain a validated tool inventory. Stage 3 normalizes tools into a unified manifest with standardized signatures, canonical arguments, and aligned output schemas. Stage 4 annotates each tool with finance attributes (timeliness, intent type, regulatory domain). Stage 5 sources and selects tool-required questions. Stage 6 aligns questions with tools via semantic retrieval, multi-sample verification, and execution checks. Stage 7 adds human-in-the-loop quality assurance. Stage 8 outputs the benchmark tool library and benchmark question set as a runnable benchmark.}
\label{fig:data_pipeline}
\end{figure*}

\subsection{Evaluation Protocols}
Because answer correctness is hard to score at scale for open-ended questions, recent work uses LLMs as judges with structured rubrics~\citep{zheng2023judging,liu2023g}, while noting that LLM-based scoring can be unstable across runs and sensitive to prompting~\citep{hashemi2024llm,haldar2025rating,d2025yescieval}, and that comparative setups can elicit more informative judgments than independent scoring~\citep{zhang2025crowd}.
In line with these findings, we reduce variance via repeated judging and explicitly separate tool execution from correctness so that a failure to call or execute tools is not conflated with an evaluation artifact; our protocol is compatible with alternative rubric designs, since the benchmark produces complete tool traces that can be inspected and re-judged.

\section{FinToolBench}
\label{sec:benchmark}

FinToolBench is an execution-grounded benchmark designed to evaluate financial tool use under real execution.
Its design emphasizes two principles.
First, every run produces an auditable tool trace.
Second, evaluation separates capability (\textit{i.e.}, invocation and execution success) from compliance (\textit{i.e.}, call-level timeliness, intent, and domain alignment).
The benchmark measures an agent's ability to select tools from a large heterogeneous library, instantiate valid arguments, handle execution failures, and produce answers whose tool use respects finance-specific constraints.

In contrast to prior tool-use benchmarks that focus primarily on API calling accuracy, FinToolBench evaluates both capability and compliance directly from executable tool traces.
Figure~\ref{fig:data_pipeline} summarizes the construction pipeline.
We first build a validated tool inventory from free-tier sources, normalize heterogeneous interfaces into a unified manifest, and annotate finance attributes for each tool.
We then construct a tool-required question set, align questions with candidate tools through retrieval and verification, and apply human-in-the-loop quality checks before release.
This design follows lines of work that stress end-to-end tool selection, argument construction, and trace-based diagnosis under real execution.

\subsection{Tool Inventory}
\label{subsec:tool_inventory}

\subsubsection{Tool Sources and Executability Filtering}
\label{subsec:executability}

We construct the tool inventory from two complementary free-tier ecosystems, ensuring reproducibility without proprietary data contracts.
\textbf{RapidAPI} is a large marketplace of third-party APIs providing broad coverage of real-time and web-based services under free-tier API keys; we filter raw endpoints with a rule-based pipeline that retains a tool only if it satisfies all of: (i) \emph{interface validity} (complete parameter definitions and non-empty descriptions); (ii) \emph{deduplication} of duplicate names and semantically identical interfaces; (iii) \emph{rate-limit sufficiency} (at least 10/h, 100/d, 300/m); (iv) \emph{authentication feasibility} under free-tier access; and (v) \emph{runtime executability} via at least one successful test invocation. Endpoints with broken URLs, faulty authentication flows, or persistent failures are discarded.
\textbf{AkShare} is an open-source Python library offering stable, research-oriented interfaces over a wide range of financial domains; we select interfaces using finance-related function-name cues (\textit{e.g.}, stock, fund, bond, futures, option, index, macro, currency, crypto, rate, treasury, ETF) and verify executability through direct invocation.

\paragraph{Scale.}
We start from 5{,}470 candidate interfaces (4{,}507 RapidAPI endpoints and 963 AkShare interfaces).
After the above filtering, the final tool library contains 760 tools.
Full criteria and counts are given in Appendix~\ref{sec:tool_curation}.

\subsubsection{Tool Normalization and Manifest Construction}
\label{subsec:normalization}

To make the heterogeneous tool ecosystem amenable to retrieval,
planning, and evaluation, we normalize each tool
into a unified manifest schema.
Each tool manifest includes:
(i) a stable identifier,
(ii) a short description,
(iii) a machine-readable signature with canonicalized parameter names and types.
Normalization reduces avoidable agent errors: date and time fields follow consistent formats,
common identifiers (\textit{e.g.}, tickers) document explicit market conventions,
and output schemas are aligned across sources.

\paragraph{Tool traces.}
Every tool invocation is captured as a structured execution trace, the atomic unit of auditing, error diagnosis, and compliance evaluation. Each record preserves step index, \texttt{tool\_name}, JSON arguments, raw output, and execution error, enabling reconstruction of the agent's reasoning chain and separation of model reasoning errors from system-level failures. The full schema is given in Appendix~\ref{sec:trace_schema}.

\begin{table*}[t]
\centering
\caption{Finance attribute schema used in FinToolBench.}
\label{tab:attrs}
\small
\begin{tabular*}{\textwidth}{@{}@{\extracolsep{\fill}}p{0.24\textwidth}p{0.30\textwidth}p{0.38\textwidth}@{}}
\toprule
Attribute & Values & Evaluation role \\
\midrule
\texttt{timeliness} &
\realt, \dailyt, \asfiledt, \periodict, \statict &
Penalize stale calls when timeliness is required. \\
\texttt{intent\_type} &
\informational, \advisory, \transactional &
Penalize escalation beyond user intent. \\
\texttt{regulatory\_domain} &
set-valued &
Penalize domain-mismatched tool usage. \\
\bottomrule
\end{tabular*}
\end{table*}

\subsubsection{Finance Attribute Annotation}
\label{subsec:finance_attributes}
Financial constraints are frequently implicit within user queries, rendering compliance measurement impossible based on raw execution traces alone. To bridge this gap, \bench{} incorporates a lightweight finance attribute schema that explicitly annotates every tool in the library. The structured metadata enables both the baseline methods outlined in Section~\ref{sec:method} and the quantitative evaluation metrics in Eq.~\eqref{eq:mismatch} to rigorously assess operational acceptability.

As summarized in Table~\ref{tab:attrs}, each tool is categorized along three distinct dimensions. These annotations are generated through an LLM-based labeling pipeline utilizing a three-vote majority agreement protocol to ensure consistency. Comprehensive details regarding the labeling rubric are provided in Appendix~\ref{sec:attributes}. By embedding these constraints directly into the tool definitions, our design decouples compliance standards from the agent under test, facilitating precise, trace-level auditing of domain mismatches.

\subsection{Question Set Construction}
\label{subsec:questions}

\paragraph{Sources and selection.}
Tool-required questions are adapted from existing finance QA datasets, including FinanceBench~\citep{islam2023financebench} and OpenFinData (\texttt{openfindata\_}\allowbreak\texttt{release})~\citep{openfindata2024}. We standardize all sources into a unified \texttt{\{question, answer, category\}} format and retain only questions identified by Qwen3-8B as requiring tool calls, with length capped at 500 characters. To ensure \bench{} strictly evaluates external tool use rather than parametric memory, queries answerable via static knowledge are excluded; we keep only items that need real-time market data, specific regulatory filings, or quantitative calculations.

\paragraph{Tool--question alignment.}
For each question, we first retrieve the top-20 candidate tools using BGE-M3 dense embeddings~\citep{chen2024bge}, then refine via an LLM verification step with Qwen3-8B under three-sample majority voting (kept if at least two votes). To prevent dominance by high-frequency tools, single-tool questions are grouped by tool name and capped at two random samples per tool; multi-tool questions are fully retained to preserve agentic-workflow diversity.

\paragraph{Human-in-the-loop verification.}
We complement automated alignment with a stratified spot-check by domain experts, confirming the logical necessity of the aligned tools, the consistency of the attribute annotations, and compliance with execution assumptions and output formatting. Further details are in Appendix~\ref{sec:questions}.

\subsection{Final Benchmark and Evaluation Protocol}
\label{subsec:final_benchmark}

The final benchmark comprises a unified tool library and a question set.
The tool library contains 760 tools, and the question set contains 295 questions, including 166 single-tool and 129 multi-tool.
Each evaluation run produces a final answer and a complete tool trace, enabling joint assessment of capability and finance compliance.

\subsection{Evaluation Metrics}
\label{subsec:bench_metrics}

We evaluate each run using two groups of metrics derived from the same auditable tool trace: capability, and compliance.
Capability measures whether an agent uses tools and whether tool-augmented traces execute successfully.
\tir{} (Tool Invocation Rate) is the fraction of samples with non-empty tool calls.
\tesr{} (Tool Execution Success Rate) is the fraction of samples whose tool-augmented traces execute successfully.
We mark a sample successful when its final tool call returns a valid parsed output without error or exception. Intermediate failures and retries are allowed.
\cer{} is the conditional execution success rate, defined as $\cer=\tesr/\tir$ (0 when $\tir=0$).
Answer correctness is captured by \soft{} and \css{}.
Numeric and choice tasks are scored against the gold answer with binary labels, while structured and free-text tasks are evaluated by the LLM judge (GPT-5.1) with scores in $\{1, 0.5, 0\}$, averaged across three repeats.
\css{} is the mean \soft{} over samples with successful execution.

Compliance metrics are defined over executed tool-call traces.
For each question $q$ with trace $\tau=\{(t_k,x_k,o_k)\}_{k=1}^{m}$, we look up each tool's finance tags from metadata,
$A(t)=(\tau_t(t), i(t), d(t))$ (timeliness, intent type, regulatory domains), and use an LLM judge (GPT-5.1) to assess
per-call alignment in each dimension. We then mark a question as mismatched if \emph{any} call in its trace is judged
to violate the corresponding constraint:
\begin{equation}
\label{eq:mismatch}
\begin{aligned}
TMF(q,\tau) &= \mathbf{1}\!\left[\exists k:\ J_T\!\big(q,A(t_k),\tau_k\big)=0\right],\\
IMF(q,\tau) &= \mathbf{1}\!\left[\exists k:\ J_I\!\big(q,A(t_k),\tau_k\big)=0\right],\\
DMF(q,\tau) &= \mathbf{1}\!\left[\exists k:\ J_D\!\big(q,A(t_k),\tau_k\big)=0\right].
\end{aligned}
\end{equation}
We then compute the dataset-level mismatch rates, denoted as \tmr{}, \imr{}, and \dmr{}, by averaging $TMF$, $IMF$, and $DMF$ over all questions with at least one tool call.
Full metric definitions are given in Appendix~\ref{sec:metrics}.

\section{Finance-Aware Tool Routing (FATR)}\label{sec:method}

We provide FATR, a reference baseline that makes finance constraints explicit to a generic LLM planner.
Rather than training a specialized policy, we reshape the context given to the planner and wrap execution with stability utilities, keeping the approach implementation-friendly and model-agnostic.
FATR serves primarily as an evaluation baseline and a reference implementation.
FinToolBench itself is independent of FATR: the benchmark consists of the tool inventory, question set, normalized trace schema, and metrics, and any agent that consumes the tool manifest and emits the same structured traces can be evaluated.
In our experiments, FATR fixes retrieval and execution infrastructure so that the reported model differences isolate planner behavior under a common stack.

\begin{figure*}[t]
\centering
\includegraphics[width=0.92\textwidth]{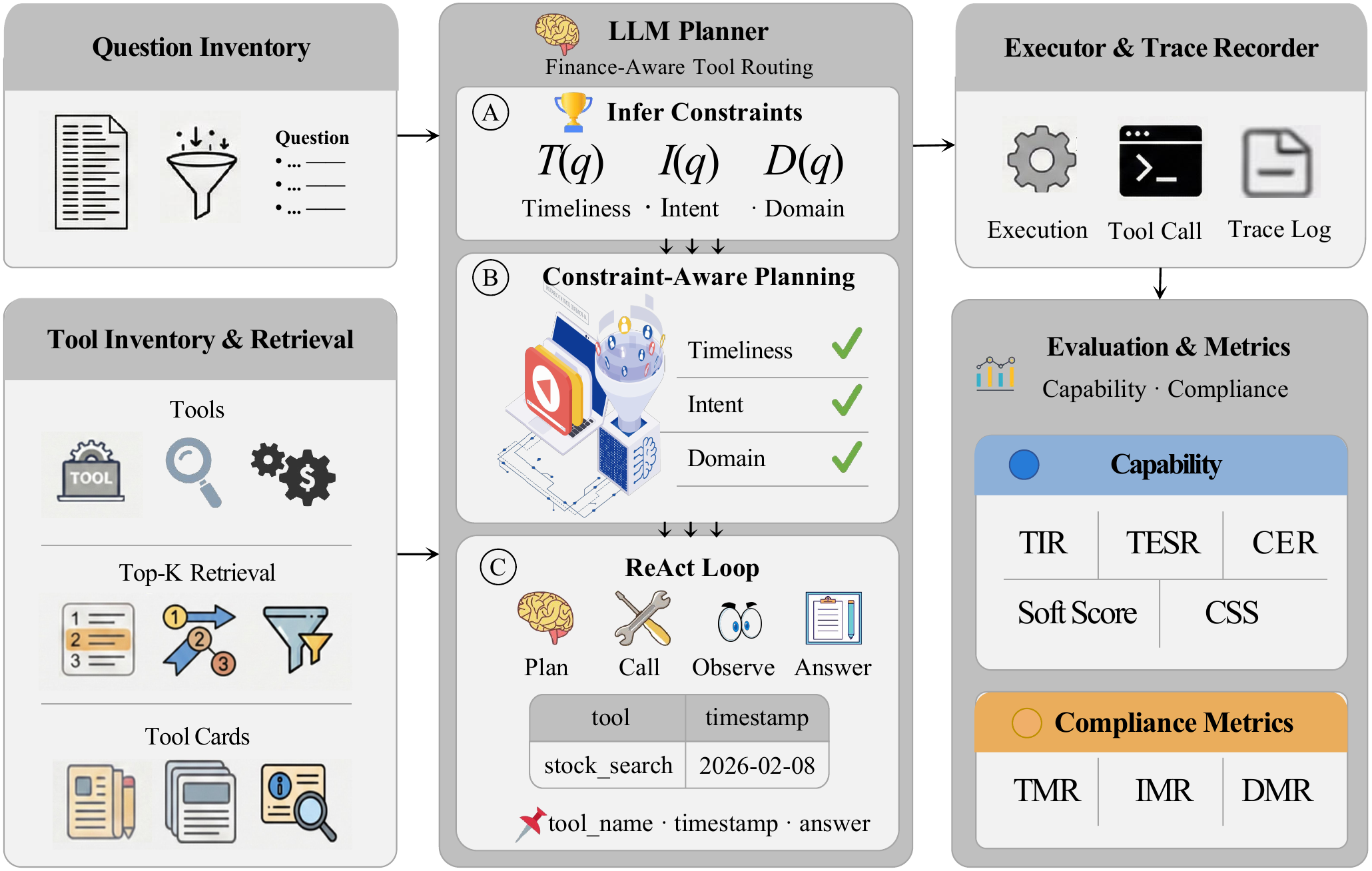}
\caption{Overview of Finance-Aware Tool Routing (FATR). FATR takes a Question Inventory and a Tool Inventory \& Retrieval module that performs Top-$K$ retrieval and formats retrieved tools as Tool Cards. An LLM Planner runs Finance-Aware Tool Routing by (A) Infer Constraints to derive $(T(q), I(q), D(q))$ over timeliness, intent, and domain, (B) Constraint-Aware Planning, and (C) a ReAct loop. An Executor \& Trace Recorder dispatches tool calls and trace logs, which are then scored by Evaluation \& Metrics for capability (\tir{}, \tesr{}, \cer{}, \soft{}, \css{}) and compliance (\tmr{}, \imr{}, \dmr{}).}
\label{fig:fatr}
\end{figure*}

\subsection{Tool Inventory \& Retrieval and Tool Cards}
Figure~\ref{fig:fatr} summarizes the FATR pipeline end to end. Each tool $t$ in the library $\mathcal{T}$ carries a callable signature and finance attributes $a_t=(t_t,i_t,d_t)$; the executor exposes all tools through a unified interface, validates structured arguments $x_k$ emitted by the planner, dispatches the call, and normalizes the returned output $o_k$ into a compact schema for caching and evaluation.
Given a question $q$ from the Question Inventory, FATR retrieves a small candidate set from Tool Inventory \& Retrieval to reduce the action space.
A retriever embeds $q$ and tool metadata and selects the Top-$K$ tools by cosine similarity (default $K{=}20$) using BGE-M3 embeddings~\citep{chen2024bge}.
Each retrieved tool is formatted as a Tool Card containing tool name and description, together with the finance attributes $a_t$.
The combination reduces distractors and makes prompting more stable under large catalogs.

\subsection{Attribute-Aware Planning}
The LLM Planner runs a ReAct loop~\citep{yao2022synergizing}: it proposes a tool call, observes tool outputs, and iterates until producing a final answer.
We cap the interaction horizon at \texttt{max\_steps{=}5} tool-augmented steps to limit latency and reduce exposure to tool drift.
During planning, FATR makes three families of constraints explicit to the planner: timeliness (\textit{e.g.}, match implied time sensitivity), intent restraint (\textit{e.g.}, avoid transactional tools unless explicitly required, and in FinToolBench transactional intent is treated as disallowed and penalized), and domain alignment (\textit{e.g.}, ensure intersection between the tool domain and the inferred question domain).
In practice, these constraints are implemented as explicit prompt rules that guide the planner's tool selection and reasoning.

\begin{figure}[t]
\centering
\includegraphics[width=\columnwidth]{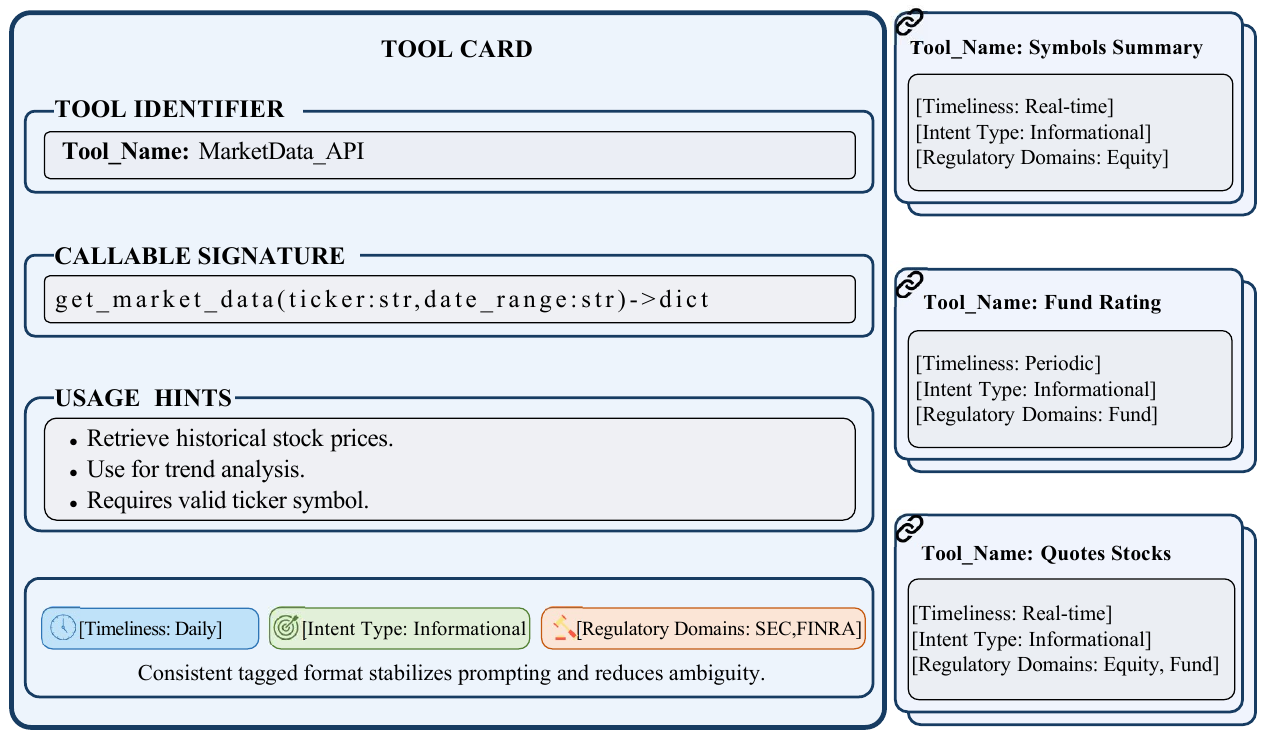}
\caption{Tool cards for attribute injection and constraint checking.}
\label{fig:toolcard}
\end{figure}

Concretely, the planner first performs Infer Constraints by articulating the implied requirement sets $(T(q), I(q), D(q))$.
It then performs Constraint-Aware Planning by selecting tools whose attributes are compatible with these sets and by maintaining the constraints throughout the ReAct loop.
For multi-tool questions, the planner is encouraged to resolve ambiguity early (\textit{e.g.}, determine the correct market and ticker format) before executing downstream calls whose domains must remain consistent.
The planner is also instructed to surface key provenance fields in the final answer so that tool use is verifiable.
Figure~\ref{fig:toolcard} illustrates our tool card format, which standardizes tool metadata and exposes finance-specific attributes used for both retrieval and constraint checking.
While the primary goal is to evaluate models rather than to enforce policy, FATR can optionally apply conservative hard filters at inference time (\textit{e.g.}, excluding transactional tools and removing domain-incompatible tools) to reduce unforced compliance errors.
Pseudocode for the full pipeline is given in Appendix~\ref{sec:algorithm}.

\section{Experiments}\label{sec:experiments}

\subsection{Benchmark Setting}
The following protocol is designed so that results on FinToolBench can be reproduced and compared fairly across studies.
The benchmark contains 295 tool-required questions and 760 runnable financial tools.
Each run uses a fixed tool-use limit, \textit{i.e.}, at most 5 tool-use rounds per question. In each round, multiple tool calls may be issued, with a per-call timeout of 60 seconds and up to 2 retries.
Tool execution is performed in a controlled environment with deterministic caching and full logging of tool traces.
For analysis, questions are stratified by single vs. multi-tool usage and by inferred question category.
We report metrics following the definitions in Section~\ref{subsec:bench_metrics}.
All headline results are computed on the full 295-question benchmark under the same fixed evaluation protocol.

\subsection{Baselines and Model Backends}
We evaluate FinToolBench under a unified agent framework based on FATR, which integrates tool retrieval, finance-attribute injection, and stabilized execution. Unless otherwise specified, the pipeline (retriever and executor) is fixed and only the LLM planner varies across seven backends: Doubao-Seed-1.6, Qwen3-8B, GLM-4.7-Flash, Claude-Sonnet-4.5, GPT-5.4, Grok-3-beta, and Gemini-3.1-Pro (preview).
All models share the same tool-call interface: the planner outputs either a final answer or a structured tool invocation specifying a tool ID and JSON arguments derived from the tool signature; outputs are returned in a normalized format, and retrieved tools are converted into function schemas with the finance tags (timeliness, intent type, regulatory domains) prepended to each tool description. The prompt emphasizes (i) using tools when timeliness is required, (ii) avoiding transactional actions, and (iii) explicitly checking that the selected tools' domains match the question. When tool outputs exceed a length threshold, an LLM-based extractor compresses responses to question-relevant fields before returning them to the planner.
For answer correctness and requirement inference, we employ an LLM-as-a-judge.
Implementation details, prompt templates, and compression settings are reported in Appendix~\ref{sec:commands} and Appendix~\ref{sec:prompt_templates}.

\subsection{Judges and Agreement}
For computing \soft{}, we use GPT-5.1 as the judge and repeat each judgment three times to reduce variance, averaging the three judge scores.
For compliance evaluation in Eq.~\eqref{eq:mismatch}, the current implementation uses GPT-5.1 and performs one LLM-judge decision per tool call for each mismatch dimension.
We report \tmr{}, \imr{}, and \dmr{} conditioned on traces with at least one tool call; therefore these mismatch rates should be interpreted together with \tir{} and \tesr{}, since conservative models can obtain low mismatch rates by attempting fewer tool-required questions.
To validate the LLM-based pipeline, three finance-domain experts independently annotate a stratified sample of 60 questions and 50 traces.
Inter-expert agreement is 85.3\%, and expert-vs-pipeline agreement is 83.8\% with F1 0.81 after second-round disagreement review.
Additional robustness checks, including prompt-order sensitivity and a Qwen3-8B compliance re-judge, are reported in Appendix~\ref{sec:robustness}.

\section{Results}
\label{sec:results}

\begin{table*}[htbp]
\centering
\caption{Main results on FinToolBench. Higher is better for \tir{}, \tesr{}, \cer{}, \soft{}, and \css{}; lower is better for \tmr{}, \imr{}, and \dmr{}.}
\label{tab:main_results}
\small
\scriptsize
\begin{tabular*}{\textwidth}{@{}@{\extracolsep{\fill}}lcccccccc@{}}
\toprule
Model & \tir & \tesr & \cer$\uparrow$ & \soft$\uparrow$ & \css$\uparrow$ & \tmr$\downarrow$ & \imr$\downarrow$ & \dmr$\downarrow$ \\
\midrule
Doubao-Seed-1.6            & 0.6508 & 0.3254 & 0.5000 & 0.4627 & 0.3958 & 0.3438 & 0.6563 & 0.1719 \\
Qwen3-8B                   & 0.8712 & 0.2949 & 0.3385 & 0.4040 & 0.4234 & 0.3307 & 0.6887 & 0.1673 \\
GLM-4.7-Flash              & 0.4407 & 0.2102 & 0.4769 & 0.3791 & 0.2769 & 0.4615 & 0.7231 & 0.1769 \\
Claude-Sonnet-4.5          & 0.4814 & 0.2407 & 0.5000 & 0.7119 & 0.5775 & 0.2746 & 0.6479 & 0.0423 \\
GPT-5.4                    & 0.2407 & 0.1322 & 0.5493 & 0.6254 & 0.4103 & 0.2535 & 0.5775 & 0.0141 \\
Grok-3-beta                & 0.2576 & 0.1932 & 0.7500 & 0.6847 & 0.4298 & 0.1316 & 0.4868 & 0.0132 \\
Gemini-3.1-Pro (preview)   & 0.1525 & 0.1186 & 0.7778 & 0.8220 & 0.5857 & 0.1333 & 0.2667 & 0.0444 \\
\bottomrule
\end{tabular*}
\end{table*}

\subsection{Main Results}
Table~\ref{tab:main_results} reports the main results on FinToolBench.
\textbf{Tool invocation does not imply execution success.}
Qwen3-8B invokes tools most often (\tir=0.8712), but its lower \cer{} shows that aggressive tool use does not necessarily translate into reliable execution.
Doubao-Seed-1.6 achieves the best end-to-end execution success (\tesr=0.3254), suggesting a stronger balance between attempting tool-required questions and completing executions successfully.
\textbf{Answer quality and trace compliance diverge.}
Claude-Sonnet-4.5 and Gemini-3.1-Pro obtain stronger semantic scores, while Grok-3-beta is the most precise among attempted tool traces with the highest \cer{} and low mismatch rates.
GPT-5.4 shows conservative tool use with moderate conditional precision, and GLM-4.7-Flash is weaker across most capability and quality metrics.
Together, these patterns show that FinToolBench separates coverage, execution reliability, answer quality, and finance-specific compliance rather than collapsing them into a single success rate.
They also indicate that model ranking depends on the operational objective: a research assistant may favor higher semantic quality, whereas a monitoring workflow may prioritize low mismatch rates and auditable traces.

\begin{figure}[t]
\centering
\includegraphics[width=\columnwidth]{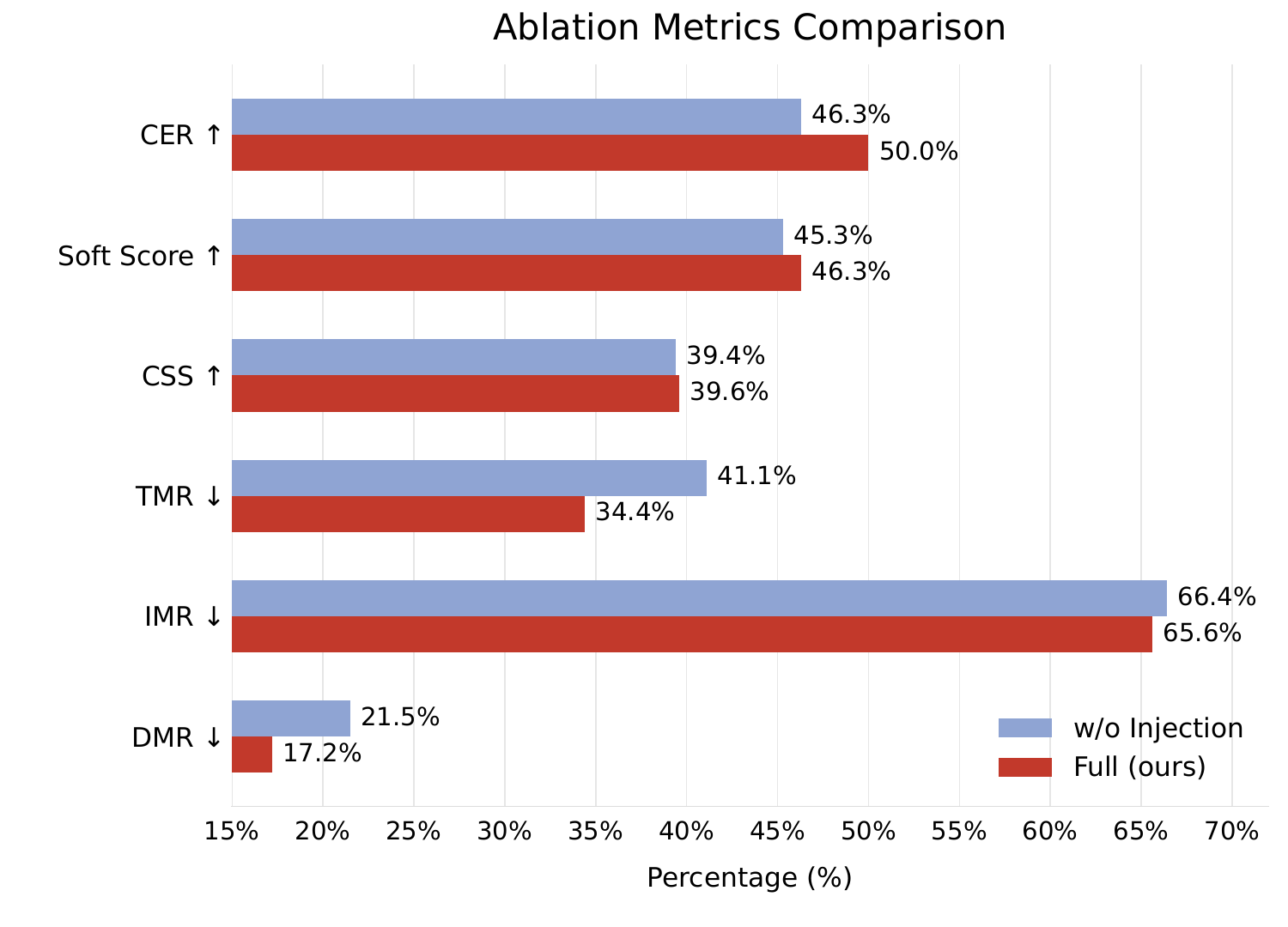}
\caption{Attribute injection ablation in \methodname{}. We compare full tool cards with finance tags against a variant without attribute injection. Attribute injection improves execution success conditioned on tool use (\cer) and reduces mismatch rates (\tmr{}, \imr{}, \dmr{}).}
\label{fig:ablation_injection}
\end{figure}

\subsection{Finance Attribute Injection}
We compare full \methodname{} against a variant that omits finance tags from the tool cards while keeping the same retriever and executor. We perform this ablation using Doubao-Seed-1.6 as the planner.
The no-injection baseline obtains \tir=0.7254, \tesr=0.3356, \cer=0.4626, \soft=0.4530, \css=0.3940, \tmr=0.4110, \imr=0.6640, and \dmr=0.2150.
Figure~\ref{fig:ablation_injection} shows that attribute injection mainly changes selection behavior: it can reduce marginal calls, improve conditional execution reliability, and reduce mismatch rates across timeliness, intent, and domain dimensions.
This supports the role of finance attributes as routing constraints rather than merely descriptive metadata.
The effect is most relevant when a question underspecifies whether the answer needs current market data, historical filings, or policy-level sources, because attribute tags help rule out plausible but misaligned tools before execution.
In practice, this makes the tool card closer to a financial interface contract: the planner must match not only argument types, but also the evidential role and regulatory scope of the call.

\begin{figure}[t]
\centering
\includegraphics[width=.84\columnwidth]{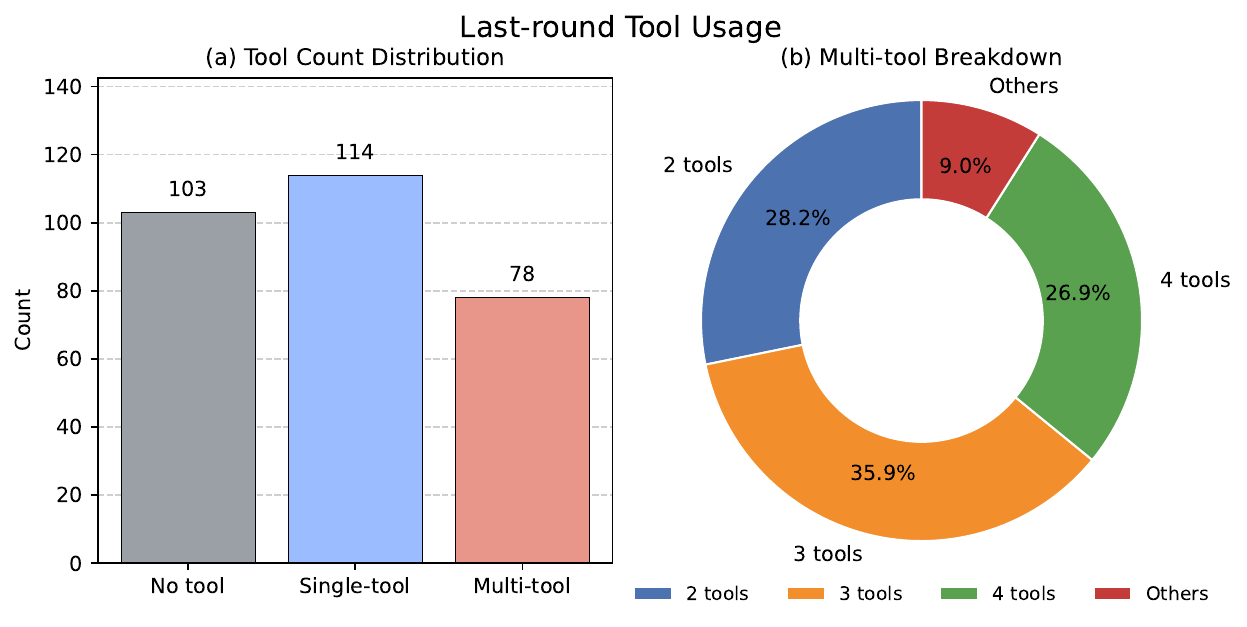}
\caption{Last-round tool usage on FinToolBench.}
\label{fig:toolcount_last_round}
\end{figure}

\subsection{Tool Usage Distribution}
Figure~\ref{fig:toolcount_last_round} shows that 103/114/78 of the 295 evaluation instances end with no/single/multiple final-round tool calls.
Doubao-Seed-1.6 makes 923 tool calls and exercises 236 distinct tools; the gold required-tool distribution is 1/2/3/4/5/6 tools = 166/64/49/12/3/1.
Thus, the 760-tool inventory functions as a realistic routing space, not as a per-tool unit-test suite.
The long-tail usage pattern suggests that stronger agents should combine broad retrieval with restraint, using multi-call traces mainly when evidence aggregation is required.

\subsection{Category-Level Diagnosis}
Appendix~\ref{sec:category_diagnosis} further breaks down capability and compliance by category, showing that aggregate scores can hide failures specific to value extraction, macro interpretation, and other tool scopes.
This diagnosis helps distinguish tool-selection failures from rigid output-format or answer-scoring mismatches.
This is important in finance, where superficially similar questions may require different data freshness, tool domains, and compliance constraints.

\section{Conclusion}
We present FinToolBench, a runnable benchmark for financial tool-use agents with 760 free-tier tools and 295 tool-required questions.
Its evaluation separates capability metrics from compliance metrics over timeliness, intent, and domain constraints.
Across seven LLM backends, no model dominates both sides: aggressive callers improve coverage but introduce noisy traces, while conservative models produce cleaner but sparser tool use.
We also provide FATR as a reference baseline and release the tool manifest, question set, and evaluation scripts to support reproducible comparison.
Future work may extend FinToolBench to paid real-time data feeds, richer market coverage, and stronger policy-constrained agents, while preserving the trace-level auditability that financial tool use requires.

\section*{Limitations}
FinToolBench covers 760 free-tier tools and 295 tool-required questions, not paid terminals, brokerage or order-routing systems, or all markets and jurisdictions.
Its metrics are benchmark checks, not professional compliance review.
API, data, and model drift, together with residual LLM-judge sensitivity, mean results should be read as a dated protocol snapshot; FATR is diagnostic, not deployable.

\section*{Ethical Considerations}
\bench{} is a research benchmark, not financial advice or authorization for regulated automation.
It uses read-only or sandboxed tools, disallows transactions, excludes personal data, credentials, paid-feed content, and non-public financial information, and any production use would require license compliance, privacy protection, human oversight, security validation, and review.


\bibliography{references}

@article{yao2022synergizing,
  title={Synergizing reasoning and acting in language models},
  author={Yao, S and Zhao, J and Yu, D and Du, N and Shafran, I and Narasimhan, K and Cao, YR},
  journal={arXiv preprint arXiv:2210.03629},
  year={2022}
}

@article{schick2023toolformer,
  title={Toolformer: Language models can teach themselves to use tools},
  author={Schick, Timo and Dwivedi-Yu, Jane and Dess{\`\i}, Roberto and Raileanu, Roberta and Lomeli, Maria and Hambro, Eric and Zettlemoyer, Luke and Cancedda, Nicola and Scialom, Thomas},
  journal={Advances in neural information processing systems},
  volume={36},
  pages={68539--68551},
  year={2023}
}

@inproceedings{li2023api,
  title={Api-bank: A comprehensive benchmark for tool-augmented llms},
  author={Li, Minghao and Zhao, Yingxiu and Yu, Bowen and Song, Feifan and Li, Hangyu and Yu, Haiyang and Li, Zhoujun and Huang, Fei and Li, Yongbin},
  booktitle={Proceedings of the 2023 conference on empirical methods in natural language processing},
  pages={3102--3116},
  year={2023}
}

@inproceedings{qin2024toolllm,
  title={Toolllm: Facilitating large language models to master 16000+ real-world apis},
  author={Qin, Yujia and Liang, Shihao and Ye, Yining and Zhu, Kunlun and Yan, Lan and Lu, Yaxi and Lin, Yankai and Cong, Xin and Tang, Xiangru and Qian, Bill and others},
  booktitle={International Conference on Learning Representations},
  volume={2024},
  pages={9695--9717},
  year={2024}
}

@inproceedings{guo2024stabletoolbench,
  title={Stabletoolbench: Towards stable large-scale benchmarking on tool learning of large language models},
  author={Guo, Zhicheng and Cheng, Sijie and Wang, Hao and Liang, Shihao and Qin, Yujia and Li, Peng and Liu, Zhiyuan and Sun, Maosong and Liu, Yang},
  booktitle={Findings of the Association for Computational Linguistics: ACL 2024},
  pages={11143--11156},
  year={2024}
}

@article{islam2023financebench,
  title={Financebench: A new benchmark for financial question answering},
  author={Islam, Pranab and Kannappan, Anand and Kiela, Douwe and Qian, Rebecca and Scherrer, Nino and Vidgen, Bertie},
  journal={arXiv preprint arXiv:2311.11944},
  year={2023}
}

@misc{openfindata2024,
  title = {{OpenFinData}: An Open Financial Evaluation Dataset},
  author = {{OpenCompass}},
  year = {2024},
  url = {https://github.com/open-compass/OpenFinData},
  note = {GitHub repository}
}

@article{zheng2023judging,
  title={Judging llm-as-a-judge with mt-bench and chatbot arena},
  author={Zheng, Lianmin and Chiang, Wei-Lin and Sheng, Ying and Zhuang, Siyuan and Wu, Zhanghao and Zhuang, Yonghao and Lin, Zi and Li, Zhuohan and Li, Dacheng and Xing, Eric and others},
  journal={Advances in neural information processing systems},
  volume={36},
  pages={46595--46623},
  year={2023}
}

@inproceedings{liu2023g,
  title={G-eval: NLG evaluation using gpt-4 with better human alignment},
  author={Liu, Yang and Iter, Dan and Xu, Yichong and Wang, Shuohang and Xu, Ruochen and Zhu, Chenguang},
  booktitle={Proceedings of the 2023 conference on empirical methods in natural language processing},
  pages={2511--2522},
  year={2023}
}

@article{chen2024bge,
  title={Bge m3-embedding: Multi-lingual, multi-functionality, multi-granularity text embeddings through self-knowledge distillation},
  author={Chen, Jianlv and Xiao, Shitao and Zhang, Peitian and Luo, Kun and Lian, Defu and Liu, Zheng},
  journal={arXiv preprint arXiv:2402.03216},
  volume={4},
  number={5},
  year={2024}
}

@article{bigeard2025finance,
  title={Finance agent benchmark: Benchmarking llms on real-world financial research tasks},
  author={Bigeard, Antoine and Nashold, Langston and Krishnan, Rayan and Wu, Shirley},
  journal={arXiv preprint arXiv:2508.00828},
  year={2025}
}

@article{lu2026beyond,
  title={Beyond Static Tools: Test-Time Tool Evolution for Scientific Reasoning},
  author={Lu, Jiaxuan and Kong, Ziyu and Wang, Yemin and Fu, Rong and Wan, Haiyuan and Yang, Cheng and Lou, Wenjie and Sun, Haoran and Wang, Lilong and Jiang, Yankai and others},
  journal={arXiv preprint arXiv:2601.07641},
  year={2026}
}

@article{jiang2025scp,
  title={SCP: Accelerating Discovery with a Global Web of Autonomous Scientific Agents},
  author={Jiang, Yankai and Lou, Wenjie and Wang, Lilong and Tang, Zhenyu and Feng, Shiyang and Lu, Jiaxuan and Sun, Haoran and Pan, Yaning and Gu, Shuang and Su, Haoyang and others},
  journal={arXiv preprint arXiv:2512.24189},
  year={2025}
}

@inproceedings{wan2026deep,
  title={Deep Research Arena: The First Exam of LLMs’ Research Abilities via Seminar-Grounded Tasks},
  author={Wan, Haiyuan and Yang, Chen and Yu, Junchi and Tu, Meiqi and Lu, Jiaxuan and Yu, Di and Cao, Jianbao and Gao, Ben and Xie, Jiaqing and Wang, Aoran and others},
  booktitle={Proceedings of the AAAI Conference on Artificial Intelligence},
  volume={40},
  number={39},
  pages={33341--33349},
  year={2026}
}

@article{yang2025multi,
  title={From what to why: A multi-agent system for evidence-based chemical reaction condition reasoning},
  author={Yang, Cheng and Lu, Jiaxuan and Wan, Haiyuan and Yu, Junchi and Qin, Feiwei},
  journal={arXiv preprint arXiv:2509.23768},
  year={2025}
}

@inproceedings{guo2025fineval,
  title={Fineval: A chinese financial domain knowledge evaluation benchmark for large language models},
  author={Guo, Xin and Xia, Haotian and Liu, Zhaowei and Cao, Hanyang and Yang, Zhi and Liu, Zhiqiang and Wang, Sizhe and Niu, Jinyi and Wang, Chuqi and Wang, Yanhui and others},
  booktitle={Proceedings of the 2025 Conference of the Nations of the Americas Chapter of the Association for Computational Linguistics: Human Language Technologies (Volume 1: Long Papers)},
  pages={6258--6292},
  year={2025}
}

@article{guo2025flame,
  title={FLAME: Financial large-language model assessment and metrics evaluation},
  author={Guo, Jiayu and Guo, Yu and Li, Martha and Tan, Songtao},
  journal={arXiv preprint arXiv:2501.06211},
  year={2025}
}

@article{patil2024gorilla,
  title={Gorilla: Large language model connected with massive apis},
  author={Patil, Shishir G and Zhang, Tianjun and Wang, Xin and Gonzalez, Joseph E},
  journal={Advances in Neural Information Processing Systems},
  volume={37},
  pages={126544--126565},
  year={2024}
}

@inproceedings{liu2024agentbench,
  title={Agentbench: Evaluating llms as agents},
  author={Liu, Xiao and Yu, Hao and Zhang, Hanchen and Xu, Yifan and Lei, Xuanyu and Lai, Hanyu and Gu, Yu and Ding, Hangliang and Men, Kaiwen and Yang, Kejuan and others},
  booktitle={International Conference on Learning Representations},
  volume={2024},
  pages={52989--53046},
  year={2024}
}

@inproceedings{zhou2024webarena,
  title={Webarena: A realistic web environment for building autonomous agents},
  author={Zhou, Shuyan and Xu, Frank F and Zhu, Hao and Zhou, Xuhui and Lo, Robert and Sridhar, Abishek and Cheng, Xianyi and Ou, Tianyue and Bisk, Yonatan and Fried, Daniel and others},
  booktitle={International Conference on Learning Representations},
  volume={2024},
  pages={15585--15606},
  year={2024}
}

@inproceedings{chen2021finqa,
  title={Finqa: A dataset of numerical reasoning over financial data},
  author={Chen, Zhiyu and Chen, Wenhu and Smiley, Charese and Shah, Sameena and Borova, Iana and Langdon, Dylan and Moussa, Reema and Beane, Matt and Huang, Ting-Hao and Routledge, Bryan R and others},
  booktitle={Proceedings of the 2021 Conference on Empirical Methods in Natural Language Processing},
  pages={3697--3711},
  year={2021}
}

@inproceedings{zhu2021tat,
  title={TAT-QA: A question answering benchmark on a hybrid of tabular and textual content in finance},
  author={Zhu, Fengbin and Lei, Wenqiang and Huang, Youcheng and Wang, Chao and Zhang, Shuo and Lv, Jiancheng and Feng, Fuli and Chua, Tat-Seng},
  booktitle={Proceedings of the 59th annual meeting of the Association for Computational Linguistics and the 11th international joint conference on natural language processing (volume 1: long papers)},
  pages={3277--3287},
  year={2021}
}

@inproceedings{patil2025berkeley,
  title={The berkeley function calling leaderboard (bfcl): From tool use to agentic evaluation of large language models},
  author={Patil, Shishir G and Mao, Huanzhi and Yan, Fanjia and Ji, Charlie Cheng-Jie and Suresh, Vishnu and Stoica, Ion and Gonzalez, Joseph E},
  booktitle={Forty-second International Conference on Machine Learning},
  year={2025}
}

@inproceedings{mialon2024gaia,
  title={Gaia: a benchmark for general ai assistants},
  author={Mialon, Gr{\'e}goire and Fourrier, Cl{\'e}mentine and Wolf, Thomas and LeCun, Yann and Scialom, Thomas},
  booktitle={International Conference on Learning Representations},
  volume={2024},
  pages={9025--9049},
  year={2024}
}

@article{drouin2024workarena,
  title={Workarena: How capable are web agents at solving common knowledge work tasks?},
  author={Drouin, Alexandre and Gasse, Maxime and Caccia, Massimo and Laradji, Issam H and Del Verme, Manuel and Marty, Tom and Boisvert, L{\'e}o and Thakkar, Megh and Cappart, Quentin and Vazquez, David and others},
  journal={arXiv preprint arXiv:2403.07718},
  year={2024}
}

@inproceedings{hashemi2024llm,
  title={Llm-rubric: A multidimensional, calibrated approach to automated evaluation of natural language texts},
  author={Hashemi, Helia and Eisner, Jason and Rosset, Corby and Van Durme, Benjamin and Kedzie, Chris},
  booktitle={Proceedings of the 62nd Annual Meeting of the Association for Computational Linguistics (Volume 1: Long Papers)},
  pages={13806--13834},
  year={2024}
}

@article{haldar2025rating,
  title={Rating Roulette: Self-Inconsistency in LLM-As-A-Judge Frameworks},
  author={Haldar, Rajarshi and Hockenmaier, Julia},
  journal={arXiv preprint arXiv:2510.27106},
  year={2025}
}

@inproceedings{d2025yescieval,
  title={Yescieval: Robust llm-as-a-judge for scientific question answering},
  author={D’Souza, Jennifer and Giglou, Hamed Babaei and M{\"u}nch, Quentin},
  booktitle={Proceedings of the 63rd Annual Meeting of the Association for Computational Linguistics (Volume 1: Long Papers)},
  pages={13749--13783},
  year={2025}
}

@inproceedings{zhang2025crowd,
  title={Crowd comparative reasoning: Unlocking comprehensive evaluations for LLM-as-a-judge},
  author={Zhang, Qiyuan and Wang, Yufei and Jiang, Yuxin and Li, Liangyou and Wu, Chuhan and Wang, Yasheng and Jiang, Xin and Shang, Lifeng and Tang, Ruiming and Lyu, Fuyuan and others},
  booktitle={Proceedings of the 63rd Annual Meeting of the Association for Computational Linguistics (Volume 1: Long Papers)},
  pages={5059--5074},
  year={2025}
}

@article{xia2025safetoolbench,
  title={SafeToolBench: Pioneering a Prospective Benchmark to Evaluating Tool Utilization Safety in LLMs},
  author={Xia, Hongfei and Wang, Hongru and Liu, Zeming and Yu, Qian and Guo, Yuhang and Wang, Haifeng},
  journal={arXiv preprint arXiv:2509.07315},
  year={2025}
}

@article{tur2025safearena,
  title={Safearena: Evaluating the safety of autonomous web agents},
  author={Tur, Ada Defne and Meade, Nicholas and L{\`u}, Xing Han and Zambrano, Alejandra and Patel, Arkil and Durmus, Esin and Gella, Spandana and Sta{\'n}czak, Karolina and Reddy, Siva},
  journal={arXiv preprint arXiv:2503.04957},
  year={2025}
}

@article{narasimhan2024tau,
  title={{$\tau$}-bench: A benchmark for tool-agent-user interaction in real-world domains},
  author={Yao, Shunyu and Shinn, Noah and Razavi, Pedram and Narasimhan, Karthik R},
  journal={arXiv preprint arXiv:2406.12045},
  year={2024}
}

@article{barres2025tau2,
  title={{$\tau^2$}-bench: Evaluating conversational agents in a dual-control environment},
  author={Barres, Victor and Dong, Honghua and Ray, Soham and Si, Xujie and Narasimhan, Karthik R},
  journal={arXiv preprint arXiv:2506.07982},
  year={2025}
}

\clearpage
\appendix

\makeatletter
\setlength{\@dblfptop}{0pt}
\setlength{\@dblfpsep}{8pt plus 2pt minus 2pt}
\setlength{\@dblfpbot}{0pt plus 1fil}
\makeatother

\section{Tool Curation Criteria}
\label{sec:tool_curation}

This section spells out the criteria used to build the \bench{} tool inventory.
The pipeline retains only tools that are executable under free-tier constraints.

\subsection{RapidAPI Endpoints}

Our RapidAPI pool is initialized from finance-related tools collected from the ToolBench paper, and then filtered with the two-stage pipeline below.

\paragraph{First-stage (rule-based) filters.}
Because RapidAPI listings vary in how authentication, billing, and rate limits are described, we programmatically crawl and parse endpoint pages to extract these fields, and then apply the deterministic criteria in Table~\ref{tab:rapidapi_filters}. An endpoint is excluded if it fails any row.

\begin{table}[H]
\centering
\caption{First-stage RapidAPI filter criteria.}
\label{tab:rapidapi_filters}
\small
\begin{tabular*}{\columnwidth}{@{}@{\extracolsep{\fill}}p{0.27\linewidth}p{0.67\linewidth}@{}}
\toprule
Criterion & Rule \\
\midrule
Description & Missing or empty $\rightarrow$ excluded. \\
Deduplication & Duplicate names: keep first only. \\
Authentication & Exclude if \texttt{Authorization} or multi-step token (beyond single API key) required. \\
Existence & Remove if endpoint does not resolve or returns persistent errors. \\
Bank-card & Exclude if free-tier requires card binding. \\
Rate limits & Require $\ge$10/h, $\ge$100/d, $\ge$300/m. \\
Map & Use an LLM to align endpoint parameter names with the tool signature and then manually spot-check the mappings. \\
\bottomrule
\end{tabular*}
\end{table}

\paragraph{Second-stage (mapping and executability).}
Each endpoint that passes the first stage is mapped into our normalized schema.
We align its parameter names to our tool signature with an LLM and then manually spot-check the mappings.
We then test each mapped endpoint with at least one successful invocation (valid request and parsed response).
Endpoints that fail consistently (timeouts, validation errors, or empty responses) are dropped.

\paragraph{Outcome.}
After the two-stage filtering process, we obtain 261 executable RapidAPI endpoints.
All retained endpoints have at least one documented required or optional parameter.
Endpoints that are parameter-free or rely solely on implicit path/query conventions are excluded.

\subsection{AkShare Interfaces}

AkShare functions are selected and then validated for executability.

\paragraph{Finance-domain filter.}
Function names are matched against a fixed set of finance-related keywords.
A function is retained if its name (or module path) contains any of the following tokens:
\begin{itemize}[leftmargin=*,nosep]
\item \texttt{stock}, \texttt{fund}, \texttt{bond}, \texttt{futures}, \texttt{option}, \texttt{index}, \texttt{macro}, \texttt{currency}, \texttt{fx}, \texttt{crypto}, \texttt{rate}, \texttt{treasury}, \texttt{etf}
\item \texttt{finance}, \texttt{bank}, \texttt{insurance}, \texttt{security}, \texttt{derivative}, \texttt{swap}, \texttt{gold}, \texttt{commodity}, \texttt{interest}, \texttt{libor}, \texttt{shibor}, \texttt{exchange}, \texttt{margin}
\end{itemize}

\paragraph{Executability.}
Each candidate is invoked with minimal valid arguments, using defaults or small example values where possible.
Interfaces that raise import errors, signature errors, or runtime errors under a timeout are discarded.

\paragraph{Outcome.}
After finance-domain filtering and executability validation, we retain 499 AkShare interfaces in the final tool inventory.

\subsection{Combined Inventory}

After executability filtering, the final tool library contains 760 tools.
All tools in \bench{} are normalized into a single manifest schema, including tool name, description, and parameters.
This unified manifest enables auditable tool traces and call-level compliance metrics (TMR, IMR, DMR), where each run records the invoked tools together with their arguments and execution outcomes.

\section{Tool Trace Schema}
\label{sec:trace_schema}

Each tool invocation is captured as a structured execution trace.
The schema (Table~\ref{tab:trace_schema}) records the chronological context through a sequential \texttt{step} index, the specific \texttt{tool\_name}, the JSON-formatted \texttt{parameters} generated by the model, and both the raw \texttt{output} and any execution \texttt{error}, so as to differentiate model reasoning errors from system-level failures such as API rejections.

\begin{table}[h]
\centering
\caption{Normalized tool-trace fields.}
\label{tab:trace_schema}
\small
\begin{tabular*}{\columnwidth}{@{}@{\extracolsep{\fill}}p{0.25\columnwidth}p{0.65\columnwidth}@{}}
\toprule
Field & Description \\
\midrule
\texttt{step} & The sequential order of the call within the multi-turn process. \\
\texttt{tool\_name} & The identifier of the specific tool invoked (\textit{e.g.}, \texttt{symbols\_sec\_filings}). \\
\texttt{parameters} & The JSON-formatted arguments generated by the model. \\
\texttt{output} & The tool response, including data or structured error messages. \\
\texttt{error} & Overall execution status, which captures null or specific system-level failures. \\
\bottomrule
\end{tabular*}
\end{table}

\section{Finance Attribute Schema and Labeling}
\label{sec:attributes}

Each tool is annotated with finance attributes that make timeliness, intent, and domain constraints explicit.
These attributes support both tool cards (for planning) and compliance evaluation (TMR, IMR, DMR).

\subsection{Attributes of the Tools}

\begin{itemize}[leftmargin=*]
\item \textbf{Timeliness} (\texttt{timeliness}): one of\\
  \realt, \dailyt, \asfiledt, \periodict, or \statict.
  \begin{itemize}[nosep]
  \item \realt: intra-day, low latency, such as ticks and order book.
  \item \dailyt: updated once per trading day or batch, such as closing prices and NAV.
  \item \asfiledt: event-driven, when a regulated entity files, such as filings and announcements.
  \item \periodict: fixed calendar schedule, such as quarterly reports and GDP.
  \item \statict: rarely changes, such as identifiers and listing dates.
  \end{itemize}
\item \textbf{Intent type} (\texttt{intent\_\allowbreak type}): one of\\
  \informational, \advisory, or \transactional.
  \begin{itemize}[nosep]
  \item \informational: read-only data access and factual retrieval without recommendations or actions.
  \item \advisory: analysis or recommendation-oriented outputs that go beyond pure retrieval.
  \item \transactional: action-triggering operations such as order placement, transfer, or account-changing behavior.
  \end{itemize}
\item \textbf{Regulatory domain} (\texttt{regulatory\_\allowbreak domain}):\newline
  Subset of \{equity, bond, fund, forex, derivatives, macro,\\ economic\_\allowbreak policy, sentiment\_\allowbreak trading, esg, crypto\}. Multiple values are allowed per tool.
\end{itemize}

\subsection{Labeling Protocol}

Annotations are produced by an LLM (Qwen3-8B) given the tool name and description.
Each attribute is labeled independently with three samples, and the final label is the majority vote.
This protocol keeps labeling separate from the agent under test and makes the compliance layer auditable.

\section{Question Set Construction}
\label{sec:questions}

The question set is built so that every retained question requires tool use.
Questions answerable by static memorization or general reasoning are excluded.

\subsection{Sources}

Questions are drawn from the FinanceBench release and the OpenFinData release (dataset identifier \texttt{openfindata\_}\allowbreak\texttt{release}).
Only questions that require tool calls to answer are retained, such as current prices, time series, or structured data.

\subsection{Filters}

\begin{itemize}[leftmargin=*,nosep]
\item \textbf{Standardization:} Convert all sources into a unified \texttt{\{question, answer, category\}} format.
\item \textbf{Maximum length:} 500 characters to keep prompts within a fixed budget.
\item \textbf{Candidate retrieval:} For each question, retrieve Top-$K$ tools ($K{=}20$) using BGE-M3 embeddings.
\item \textbf{LLM tool selection:} Given the Top-$K$ tools, Qwen3-8B selects the most suitable tool(s). We sample three times and keep tools with at least two votes.
\item \textbf{Deduplication:} For single-tool questions, keep at most two questions per tool to reduce skew. Multi-tool questions are not deduplicated by tool set.
\end{itemize}

\subsection{Outcome}

The final question set contains 295 questions, spanning both simple and compositional tool use: 166 single-tool questions and 129 multi-tool questions, classified by the number of aligned required tools.

\subsection{Human Verification}

We conduct human-in-the-loop quality assurance as an artifact validation step rather than as an experiment on human subjects.
We use stratified sampling to select 60 questions and 50 execution traces, covering single-tool and multi-tool items as well as the three compliance dimensions.
Three domain experts in financial data analysis independently check benchmark items against a fixed rubric: whether the question truly requires tool use, whether the aligned tool or tool sequence is logically necessary, whether the annotated timeliness, intent type, and regulatory domain are plausible, and whether the expected answer format is compatible with the available tool outputs.
Reviewers do not provide personal data, are not evaluated as study participants, and only inspect benchmark artifacts.
Disagreements are resolved in a second-round review with the benchmark maintainers, and unresolved items are removed from the release.
The resulting inter-expert agreement is 85.3\%.
Against the automatic pipeline, expert agreement is 83.8\% with F1 0.81.
Dimension-level results are shown in Table~\ref{tab:human_validation}.

\begin{table}[h]
\centering
\caption{Human validation of question-side attributes and trace-level compliance.}
\label{tab:human_validation}
\small
\begin{tabular*}{\columnwidth}{@{}@{\extracolsep{\fill}}lcc@{}}
\toprule
Validation target & Agreement & Secondary metric \\
\midrule
Question timeliness & 85.0\% & Jaccard 0.82 \\
Question intent & 93.3\% & Jaccard 0.91 \\
Question domain & 80.0\% & Jaccard 0.76 \\
Trace TMR & 86.0\% & F1 0.83 \\
Trace IMR & 94.0\% & F1 0.91 \\
Trace DMR & 82.0\% & F1 0.78 \\
\bottomrule
\end{tabular*}
\end{table}

\section{Metric Definitions}
\label{sec:metrics}

All metrics are defined over a fixed question set and a fixed evaluation protocol (timeout, retries, caching).
We report capability (TIR, TESR, CER) and compliance (TMR, IMR, DMR) to separate whether an agent can run tools from whether its tool choices satisfy finance constraints.
Below, $N$ denotes the number of questions.

\subsection{Capability}

\begin{itemize}[leftmargin=*]
\item \textbf{Tool Invocation Rate (\tir).} Fraction of questions whose execution invokes at least one tool call:
\[\mathrm{TIR}=\frac{1}{N}\sum_{i=1}^{N}\mathbf{1}\!\left[\,|\mathrm{tool\_calls}_i|>0\,\right].\]
\item \textbf{Tool Execution Success Rate (\tesr).} Fraction of questions whose tool execution succeeds.
We mark a question successful when its final tool call returns a valid parsed output without error or exception; intermediate failures and retries are allowed:
\begin{equation*}
\resizebox{0.95\columnwidth}{!}{$\displaystyle\mathrm{TESR}=\frac{1}{N}\sum_{i=1}^{N}\mathbf{1}\!\left[\,\text{final tool call succeeds on question } i\,\right]$}
\end{equation*}
\item \textbf{Conditional Execution Rate (\cer).} Success rate among questions that invoked at least one tool call: 
  \[\mathrm{CER}=\frac{\mathrm{TESR}}{\mathrm{TIR}},\quad \mathrm{CER}=0 \text{ when } \mathrm{TIR}=0.\]
\item \textbf{Soft Score (\soft).} 
We partition questions into three types based on the gold-answer format:
(i) numeric/choice questions, identified by the presence of a \texttt{numeric} or \texttt{choices} field in the gold answer;
(ii) structured-analysis questions, identified by the presence of a \texttt{criterium} field in the gold answer;
(iii) \texttt{other} questions, all remaining non-numeric/choice/\texttt{criterium} cases.
All questions are judged by LLM (GPT-5.1). Numeric/choice questions are scored in $\{1,0\}$, while structured-analysis and \texttt{other} questions are scored in $\{1,0.5,0\}$. Scores are averaged over three judging repeats.
When a gold answer provides a machine-readable numeric value or option label, a deterministic exact-match or tolerance-match audit can be added as a complementary check; the current headline metric uses the LLM score because several questions require explanatory or structured answers.
\[\mathrm{SoftScore}=\frac{1}{3N}\sum_{i=1}^{N}\sum_{r=1}^{3} s_{i,r}.\]
\item \textbf{Conditional Soft Score (\css).} Mean \soft{} over questions with successful execution:
\begin{align*}
\mathrm{CSS} &= \frac{\sum_{i=1}^{N} e_i \left(\frac{1}{3}\sum_{r=1}^{3} s_{i,r}\right)}{\sum_{i=1}^{N} e_i},\\
\mathrm{CSS} &= 0 \quad \text{when } \sum_{i=1}^{N} e_i = 0.
\end{align*}
where $e_i = \mathbf{1}[\text{final tool call succeeds on question } i]$.
\end{itemize}

\subsection{Compliance Mismatch Rates}

For each question $q$, an LLM(GPT-5.1) judge assesses whether the tools used in the execution are aligned with the question's finance constraints in timeliness, intent type, and regulatory domain.
Let the executed tool-use trace be $\tau=\{(t_k,x_k,o_k)\}_{k=1}^{m}$, where $t_k$ is the tool name selected at step $k$, $x_k$ is the tool input (arguments), and $o_k$ is the tool output.
Tool attributes are not part of the trace; instead, for any tool $t$ we look up its finance tags from the tool metadata:
$A(t)=(\tau_t(t), i(t), d(t))$, corresponding to timeliness $\tau_t(t)$, intent type $i(t)$, and regulatory domains $d(t)$.

Define judge-level call alignment indicators:
$J_T(q,t_k,x_k,o_k,\tau_t(t_k))\in\{0,1\}$,
$J_I(q,t_k,x_k,o_k,i(t_k))\in\{0,1\}$,
and $J_D(q,t_k,x_k,o_k,d(t_k))\in\{0,1\}$,
where $1$ means matched and $0$ means mismatched.
We then define mismatch-at-least-once indicators at the question level:
{\small
\begin{align*}
\mathrm{TMF}(q,\tau) &= \mathbf{1}\!\left[\exists k:\ J_T(q,t_k,x_k,o_k,\tau_t(t_k))=0\right], \\
\mathrm{IMF}(q,\tau) &= \mathbf{1}\!\left[\exists k:\ J_I(q,t_k,x_k,o_k,i(t_k))=0\right], \\
\mathrm{DMF}(q,\tau) &= \mathbf{1}\!\left[\exists k:\ J_D(q,t_k,x_k,o_k,d(t_k))=0\right].
\end{align*}
}


TMR, IMR, and DMR are computed over questions with at least one executed tool call, and measure the fraction of traces that contain at least one mismatch in the corresponding dimension. Lower is better.
Let\[\mathcal{Q}_{\mathrm{tool}}=\{\,q_i \mid i\in\{1,\dots,N\}, |\mathrm{tool\_calls}_i|>0\,\}\]
denote the set of questions with at least one executed tool call. We then define
\begin{align*}
\mathrm{TMR} &= \frac{1}{|\mathcal{Q}_{\mathrm{tool}}|}\sum_{q_i\in\mathcal{Q}_{\mathrm{tool}}} \mathrm{TMF}(q_i,\tau_i), \\
\mathrm{IMR} &= \frac{1}{|\mathcal{Q}_{\mathrm{tool}}|}\sum_{q_i\in\mathcal{Q}_{\mathrm{tool}}} \mathrm{IMF}(q_i,\tau_i), \\
\mathrm{DMR} &= \frac{1}{|\mathcal{Q}_{\mathrm{tool}}|}\sum_{q_i\in\mathcal{Q}_{\mathrm{tool}}} \mathrm{DMF}(q_i,\tau_i).
\end{align*}

Because these rates are conditional on tool invocation, they diagnose the quality of attempted tool traces but do not by themselves reward coverage.
For this reason, the main table reports them jointly with \tir{} and \tesr{} so that coverage and attempted-trace quality are visible together.

\section{Prompt Templates}
\label{sec:prompt_templates}

This section documents the prompts for (i) intent mismatch evaluation, (ii) planner system instructions, and (iii) RapidAPI parameter alignment.

\subsection{Intent-Match Evaluator (IMR)}

We use an LLM (GPT-5.1) judge to decide whether a tool call’s \texttt{intent\_type} is compatible with the question requirement.
The judge considers only intent (informational, advisory, transactional) and ignores execution success.

\begin{promptcard}{Intent-match evaluator prompt (JSON-only output)}
You are a financial tool compliance evaluator. Determine whether the ``question requirement’’ matches the ``tool intent’’.\\
Consider only tool intent. Do not consider whether the tool call succeeds.\\
Tool intent must be one of \texttt{informational} / \texttt{advisory} / \texttt{transactional}.\\[4pt]
Question: \{question\}\\
Tool name: \{tool\_name\}\\
Tool intent (\texttt{intent\_type}): \{intent\_type\}\\
Tool description: \{description\}\\[4pt]
Output JSON only: \{"match": "yes$|$no", "reason": "within 20 words"\}
\end{promptcard}

\subsection{Planner System}

The planner is explicitly instructed to use the \texttt{financial\_tags} field inside each tool description when choosing tools.
This makes timeliness, domain, and intent cues salient at selection time.

\begin{promptcard}{Planner system}
Subject to satisfying the question, first infer the question's required finance tags:
(a) required \texttt{timeliness}, (b) required \texttt{regulatory\_domains}, and (c) allowed \texttt{intent\_type}.\\
Then select tools whose \texttt{financial\_tags} best match these requirements: require domain overlap, prefer matching timeliness, and choose the lowest-risk intent (\texttt{informational} > \texttt{advisory}; avoid \texttt{transactional} unless explicitly requested).\\
If multiple tools match, choose the one whose signature/description best fits the needed inputs/outputs.
\end{promptcard}

\subsection{RapidAPI Parameter Alignment}

RapidAPI documentation often uses parameter names that differ from those used in request examples.
To reduce argument instantiation errors, we generate normalized Python wrappers whose function parameters exactly match the keys used in the request schema, and we verify mappings with manual spot-checking (Table~\ref{tab:rapidapi_filters}, Map rule).
For security, the API key is shown as a placeholder.

\Needspace{0.50\textheight}
\begin{promptcard}{Parameter-aligned Python code generation prompt}
You are a professional Python tool function generator. Rewrite the following raw API call code into a standard reusable function. Strictly follow the rules below.\\[4pt]
1. Function name: \{func\_name\}\\
2. Function parameters:\\
\hspace*{1em}-- All input parameters: \{param\_str\}\\
\hspace*{1em}-- Additional fixed parameter: \texttt{rapidapi\_key: str = "<RAPIDAPI\_KEY>"}\\
3. The function must:\\
\hspace*{1em}-- Use \texttt{requests} to send the HTTP request\\
\hspace*{1em}-- Use \texttt{X-RapidAPI-Key} and \texttt{X-RapidAPI-Host} in headers\\
\hspace*{1em}-- Return \texttt{response.json()} if possible, otherwise return \texttt{response.text}\\
\hspace*{1em}-- Do not use \texttt{print}; always \texttt{return}\\
\hspace*{1em}-- Include a full docstring describing the function and every parameter\\
\hspace*{1em}-- Output pure Python code only\\
\hspace*{1em}-- Parameter names must exactly match the keys in the parameter dictionary (case sensitive)\\[4pt]
Description: \{desc\}\\
Parameter spec: \{parameters\}\\
URL: \{url\}\\
X-RapidAPI-Host: \{host\}
\end{promptcard}

\section{Additional Robustness Analyses}
\label{sec:robustness}

\subsection{Retrieval Depth}

We evaluate how often the gold required tool set is covered by the retriever as the number of retrieved candidates changes.
Table~\ref{tab:retrieval_depth} shows that retrieval improves sharply up to $K{=}20$ and then saturates, motivating the default $K{=}20$ used in the main experiments.

\begin{table}[H]
\centering
\caption{Retrieval-depth ablation. Tool Hit@$K$ measures whether the gold required tool set is covered by the retrieved candidate pool.}
\label{tab:retrieval_depth}
\small
\begin{tabular*}{\columnwidth}{@{}l@{\hspace{0.8em}}ccccc@{}}
\toprule
$K$ & 1 & 5 & 10 & 20 & 30 \\
\midrule
Tool Hit@$K$ & 61.2\% & 75.1\% & 83.0\% & 88.1\% & 88.3\% \\
\bottomrule
\end{tabular*}
\end{table}

\subsection{Judge Stability and Prompt Order}

For answer scoring, the three-repeat GPT-5.1 judge produces a representative \soft{} standard deviation of 0.02 and a 95\% confidence interval of approximately $\pm$0.03.
For compliance prompt order, we compare the original IMR prompt, which asks for the binary decision before the rationale, against a reversed-order variant on 50 sampled IMR cases.
The binary labels agree on 94.0\% of samples; the overall IMR differs by 1.0 percentage point (18.0\% vs. 19.0\%), with a 6.0\% sample-level flip rate.

\subsection{Alternative Compliance Judge}

To test whether compliance conclusions depend on GPT-5.1 alone, we rerun requirement inference with Qwen3-8B on the same traces.
For Doubao-Seed-1.6, Qwen3-8B gives \tmr/\imr/\dmr{} = 0.3346/0.6848/0.1673, close to GPT-5.1's 0.3438/0.6563/0.1719.
This does not remove all judge noise, but it suggests that the broad compliance pattern is not an artifact of a single judge model.

\section{Commands and Reproduction Checklist}
\label{sec:commands}

The following enables independent reproduction of the benchmark and evaluation pipeline so that results can be compared fairly across studies.

\subsection{Environment and Models}

RapidAPI keys are obtained following the official documentation. AkShare usage follows its documentation.
The retriever is BGE-M3, and the output compressor uses Qwen3-8B.
Planner backends are Doubao-Seed-1.6, Qwen3-8B, GLM-4.7-Flash, Claude-Sonnet-4.5, GPT-5.4, Grok-3-beta, and Gemini-3.1-Pro (preview).
The judge is GPT-5.1 with three repeats for \soft{} and one decision per tool call for requirement metrics in the current implementation.
The released code and configuration specify the evaluation stack, including prompts, decoding settings, timeout, retry count, maximum tool steps, cache setting, and prompt-template hash.
Trace logs can be replayed or re-judged under later model versions.

\subsection{Artifact Documentation and Licensing}

The release contains four artifact families: (i) the normalized tool manifest, (ii) the 295-question benchmark set, (iii) the evaluator and FATR reference implementation, and (iv) cached execution traces and schema snapshots used for replay.
For each tool, the manifest records source ecosystem (RapidAPI or AkShare), normalized signature, finance attributes, schema extraction date, last successful execution date, and whether a cached output is available.
RapidAPI endpoints are invoked under the endpoint providers' free-tier terms through user-supplied credentials; we do not redistribute API keys or paid-feed outputs.
AkShare interfaces are documented with the upstream package version and license metadata.
FinanceBench, OpenFinData, BGE-M3, RapidAPI, and AkShare are cited as upstream artifacts, and derivative benchmark files are distributed only under terms compatible with the source artifacts.
The intended use of the release is research evaluation of financial tool-use agents, trace auditing, and reproducibility studies.
It is not intended for live trading, client advice, transaction execution, or production compliance certification.

\subsection{PII and Sensitive-Content Audit}

Before release, we scan questions, tool manifests, prompts, and cached traces for personal identifiers, credentials, API keys, account numbers, emails, phone numbers, and non-public company information.
Secret-like strings are replaced with placeholders, and any item requiring user-specific account data is excluded.
The benchmark uses public or free-tier financial information and does not include private user portfolios, bank records, or proprietary market data.
Because live API outputs can change, users who regenerate traces are instructed to repeat the same audit before redistribution.


\section{Algorithm --- FATR Pipeline}
\label{sec:algorithm}

\begin{algorithm}[t]
\small
\caption{FATR pipeline}
\begin{algorithmic}[1]
\Require Question $q$, tool library $\mathcal{T}$, retriever $\mathcal{R}$, planner LLM $\pi$, Top-$K$ (default $K{=}20$)
\Ensure Final answer $\hat{y}$, tool trace $\tau$
\State $\mathcal{C} \gets \mathcal{R}(q, \mathcal{T}, K)$ \Comment{Retrieve Top-$K$ tools by similarity}
\State Build tool cards: for each $t \in \mathcal{C}$, format $\mathrm{Card}(t)$ with signature and finance attributes
\State $\tau \gets \emptyset$, $h \gets \emptyset$ \Comment{History of tool calls and outputs}
\Repeat
\State Planner $\pi$ proposes next action given $(q,\{\mathrm{Card}(t)\}_{t\in\mathcal{C}},h)$: tool $t$ and arguments $x$ (or final answer $\hat{y}$)
\If{action is tool call $(t, x)$}
\State Validate $t \in \mathcal{C}$
\State Execute: $o \gets \mathsf{Exec}(t, x)$ with timeout, optionally compress $o$
\State Append $(t, x, o)$ to $\tau$, update $h$ with observation
\Else
\State $\hat{y} \gets$ planner output, \textbf{break}
\EndIf
\Until{planner emits final answer or max steps}
\State \Return $\hat{y}$, $\tau$
\end{algorithmic}
\end{algorithm}

The algorithm summarizes \methodname{}: retrieval, tool-card formatting, ReAct-style planning with finance constraints, and stabilized execution with timeout, retries, cache, and optional output compression.
When compression is enabled, Qwen3-8B extracts question-relevant fields from long tool responses.

\FloatBarrier

\section{Category-Level Diagnosis}
\label{sec:category_diagnosis}

\begin{figure*}[p]
\centering
\includegraphics[width=0.92\textwidth]{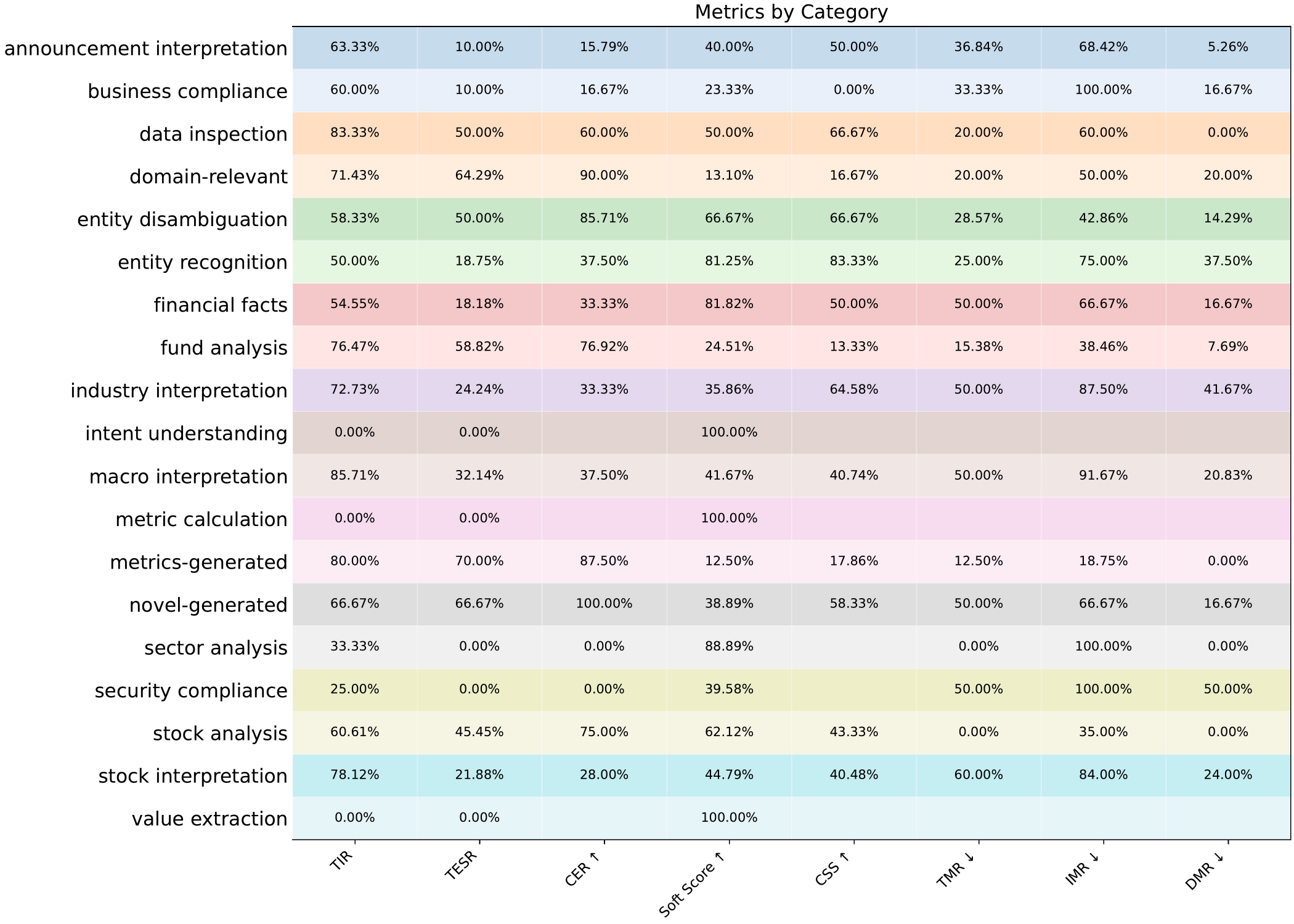}
\caption{Metrics by category for Doubao-Seed-1.6 on FinToolBench. Rows are question categories and columns are evaluation metrics.}
\Description{Heatmap of FinToolBench metrics by category for Doubao-Seed-1.6.}
\label{fig:metrics_by_category}
\end{figure*}

To further localize failure modes beyond aggregate averages, we report metrics broken down by question category.
Figure~\ref{fig:metrics_by_category} presents category-level performance for Doubao-Seed-1.6.
The heatmap highlights substantial heterogeneity across categories.
Categories with low \tir cap end-to-end execution success by limiting tool coverage.
In contrast, categories with high \tir but low \tesr indicate difficulties in argument construction, disambiguation, or endpoint instability even after the agent commits to tool use.

\section{Case Studies}
\label{sec:case_studies}

This section presents representative execution traces with condensed tool calls, final answers, correctness labels, and trace-level analysis.

\subsection{Case Study 1: Finance Attribute Injection Changes Tooling Behavior}
\label{sec:case_study_financebench_margin}

This case examines whether finance-attribute injection changes tool selection when the first attempted interface is partially incompatible with the benchmark wrapper.
We compare the no-attribute baseline against \methodname{} under the same retrieval pool and execution budget.

\begin{figure*}[p]
\centering
\small
\begin{minipage}{\textwidth}
\begin{casecard}{Case Study 1 (\texttt{financebench\_9}): injection recovers tool execution but not task framing}
\textbf{Question.} Does AMEX have an improving operating margin profile as of 2022? If operating margin is not a useful metric for a company like this, then state that and explain why.\par
\smallskip
\textbf{Ground truth.} ``Performance is not measured through operating margin.''\par
\medskip
\begin{minipage}[t]{0.49\textwidth}
\textbf{Baseline (no finance attributes).}\par
\textbf{Tool trace.}\par
Step 1: \texttt{symbols\_get\_fundamentals} $\rightarrow$ error (unexpected keyword argument \texttt{fields}).\par
\textbf{Final answer.} No final answer is produced.\par
\textbf{Correctness.} Incorrect or unscored.\par
\end{minipage}\hfill
\begin{minipage}[t]{0.49\textwidth}
\textbf{\methodname{} (with injected finance attributes).}\par
\textbf{Tool trace.}\par
Step 1: \texttt{symbols\_get\_fundamentals} $\rightarrow$ empty.\par
Step 2: \texttt{symbols\_get\_sector\_metrics} $\rightarrow$ single-period margin.\par
Step 3: \texttt{symbols\_get\_fundamentals} (\texttt{limit=3}) $\rightarrow$ empty.\par
Step 4: \texttt{symbols\_financials\_metrics} $\rightarrow$ FY22 operating income grows slower than revenue.\par
\textbf{Final answer.} AMEX’s operating margin did not improve in 2022; revenue outpaced operating income.\par
\textbf{Correctness.} Incorrect.\par
\end{minipage}
\medskip

\textbf{Analysis.} Finance attributes change tooling behavior. The injected run recovers from early API-interface failures and produces a tool-backed narrative, while the baseline fails to complete a trace. However, correctness still depends on aligning with the dataset's intended criterion. Here, the tool-backed answer discusses margin trends instead of adopting the ground-truth stance that operating margin is not the right performance metric.\par
\smallskip
\textbf{Takeaway.} Attribute injection improves execution continuity, but does not by itself guarantee semantic alignment with the ground-truth criterion.\par
\end{casecard}
\caption{Single-box, end-to-end comparison for \texttt{financebench\_9}. The baseline fails due to an interface mismatch. \methodname{} recovers a valid tool trace, but the final answer remains incorrect because it does not follow the dataset's intended reasoning criterion.}
\label{fig:case_study_financebench_margin}
\end{minipage}
\end{figure*}

\subsection{Case Study 2: Finance Attributes Reduce Redundant Tooling}
\label{sec:case_study_openfindata_fund}

This case focuses on trace efficiency: the question can be answered with one fund-specific informational endpoint, so extra calls mostly reflect planner uncertainty.

\begin{figure*}[p]
\centering
\small
\begin{minipage}{\textwidth}
\begin{casecard}{Case Study 2 (\texttt{openfindata\_release\_155}): injection prunes redundant tools while preserving correctness}
\textbf{Question.} As a financial analyst, evaluate the downside risk profile of the Tianhong Yu'e Bao Money Market Fund using the downside risk data provided below.\par
\smallskip
\textbf{Ground truth.} Tianhong Yu'ebao has a low downside risk and is among the better performers in its category.\par
\medskip
\begin{minipage}[t]{0.49\textwidth}
\textbf{Baseline (no finance attributes).}\par
\textbf{Tool trace.}\par
\texttt{fund\_individual\_analysis\_xq} $\rightarrow$ volatility metrics inconsistent with money funds.\par
\texttt{fund\_money\_fund\_info\_em} $\rightarrow$ timeout or serialization error.\par
\texttt{rate\_interbank} $\rightarrow$ execution error.\par
\texttt{fund\_overview\_em} $\rightarrow$ confirms low downside risk.\par
\textbf{Final answer.} Tianhong Yu'e Bao Money Market Fund exhibits extremely low downside risk, significantly below the peer average, making it suitable for conservative investors.\par
\textbf{Correctness.} Correct.\par
\end{minipage}\hfill
\begin{minipage}[t]{0.49\textwidth}
\textbf{\methodname{} (with injected finance attributes).}\par
\textbf{Tool trace.}\par
\texttt{fund\_overview\_em} $\rightarrow$ confirms near-zero downside risk.\par
\textbf{Final answer.} Over the past year and across multiple short-term horizons, Tianhong Yu'e Bao Money Market Fund exhibits near-zero downside risk, significantly outperforming the peer average.\par
\textbf{Correctness.} Correct.\par
\end{minipage}
\medskip

\textbf{Analysis.} Injection improves trace quality by pushing the planner toward a fund-specific, informational tool with appropriate scope. The baseline eventually reaches a suitable tool, but only after several failed or ill-suited calls. In this qualitatively straightforward item, both runs are correct, so the benefit mainly appears as a cleaner and more stable execution trace.\par
\smallskip
\textbf{Takeaway.} Both runs are correct, but the injected run reaches the answer with a shorter and cleaner trace.\par
\end{casecard}
\caption{Single-box comparison for \texttt{openfindata\_release\_155}. Injected finance attributes reduce redundant and incompatible tool calls while preserving correctness.}
\label{fig:case_study_openfindata_fund}
\end{minipage}
\end{figure*}

\subsection{Case Study 3: Finance Attributes Do Not Guarantee Numerical Fidelity}
\label{sec:case_study_financebench_cogs}

This case isolates a harder failure mode: the required numeric target is not directly exposed by the available tools, so the planner must choose between abstaining and using a proxy.

\begin{figure*}[p]
\centering
\small
\begin{minipage}{\textwidth}
\begin{casecard}{Case Study 3 (\texttt{financebench\_33}): injection changes strategy but remains incorrect}
\textbf{Question.} What is Nike's three-year average cost of goods sold as a percentage of revenue from FY2016 to FY2018?\par
\smallskip
\textbf{Ground truth.} 55.1\%.\par
\medskip
\begin{minipage}[t]{0.49\textwidth}
\textbf{Baseline (no finance attributes).}\par
\textbf{Tool trace.}\par
Income-statement tools $\rightarrow$ historical data locked or unavailable.\par
\texttt{stock\_get\_cost\_of\_revenue} $\rightarrow$ single-point value without year context.\par
\textbf{Final answer.} The requested three-year average cannot be calculated with the available data.\par
\textbf{Correctness.} Incorrect.\par
\end{minipage}\hfill
\begin{minipage}[t]{0.49\textwidth}
\textbf{\methodname{} (with injected finance attributes).}\par
\textbf{Tool trace.}\par
\texttt{symbols\_get\_sector\_metrics} $\rightarrow$ gross margin retrieved.\par
\textbf{Final answer.} Using gross margin (38.6\%), the implied average COGS ratio is approximately 61.4\%.\par
\textbf{Correctness.} Incorrect.\par
\end{minipage}
\medskip

\textbf{Analysis.} Injection pushes the agent away from unavailable historical statements toward a proxy metric. This avoids dead ends, but it does not ensure numerical fidelity. Even with a cleaner trace, the agent can still produce an incorrect estimate when the required quantity is not directly supported by the available tools.\par
\smallskip
\textbf{Takeaway.} Finance attributes can redirect the planner to a more plausible tool path, but they are not a substitute for numeric-grounding checks.\par
\end{casecard}
\caption{Single-box comparison for \texttt{financebench\_33}. Injected finance attributes change the tool strategy, but the resulting proxy-based estimate remains incorrect.}
\label{fig:case_study_financebench_cogs}
\end{minipage}
\end{figure*}
\FloatBarrier

\end{document}